\documentclass[acmtog,nonacm]{acmart}
\AtBeginDocument{%
  }

\usepackage{algorithm}
\usepackage{algorithmic}
\usepackage{xcolor} % 用于算法块中灰色的注释
\usepackage{amsmath} 
\usepackage{multirow}
\usepackage{graphicx} % 用于 \resizebox
\usepackage{booktabs}
\usepackage{xcolor}
\definecolor{keyblue}{RGB}{0,90,180}
\definecolor{keyred}{RGB}{200,40,40}
\definecolor{keypurple}{RGB}{130,60,180}

\usepackage{array}
\usepackage{booktabs}

\newif\ifshowchanges
\showchangestrue

\showchangesfalse

\ifshowchanges
  \newcommand{\rev}[1]{\textcolor{blue}{#1}}
  \newcommand{\revcolor}{\color{blue}}
\else
  \newcommand{\rev}[1]{#1}
  \newcommand{\revcolor}{}
\fi

\begin{document}
\author{Zixuan Duan}
\authornote{These authors contributed equally.}
\orcid{0009-0002-6915-8706}
\email{522024710003@smail.nju.edu.cn}
\affiliation{%
  \institution{Nanjing University, Institute of Artificial Intelligence, China Telecom (TeleAI)}
  \country{China}}

\author{Xunzhi Xiang}
\authornotemark[1]
\orcid{0009-0006-9629-0410}
\email{xbxsxp@gmail.com}
\affiliation{%
  \institution{Nanjing University, Institute of Artificial Intelligence, China Telecom (TeleAI)}
  \country{China}}

\author{Yabo Chen}
\authornotemark[1]
\orcid{0000-0003-1463-0188}
\email{chenyabo@sjtu.edu.cn}
\affiliation{%
  \institution{Institute of Artificial Intelligence, China Telecom (TeleAI)}
  \country{China}}

\author{Xin Zhang}
\orcid{0009-0005-7310-8316}
\email{25113090168@m.fudan.edu.cn}
\affiliation{%
  \institution{Fudan University, Institute of Artificial Intelligence, China Telecom (TeleAI)}
  \country{China}}

\author{Changhan Liu}
\orcid{0009-0006-3125-9308}
\email{502025710010@smail.nju.edu.cn}
\affiliation{%
  \institution{Nanjing University}
  \country{China}}

\author{Haibin Huang}
\orcid{0000-0002-7787-6428}
\email{jackiehuanghaibin@gmail.com}
\affiliation{%
  \institution{Institute of Artificial Intelligence, China Telecom (TeleAI)}
  \country{China}}

\author{Chi Zhang}
\orcid{0009-0002-3514-2490}
\email{zhangc120@chinatelecom.cn}
\affiliation{%
  \institution{Institute of Artificial Intelligence, China Telecom (TeleAI)}
  \country{China}}

\author{Qi Fan}
\authornote{Corresponding authors.}
\orcid{0000-0002-2644-4457}
\email{fanqi@nju.edu.cn}
\affiliation{%
  \institution{Nanjing University}
  \country{China}}

\author{Xuelong Li}
\authornotemark[2]
\orcid{0000-0002-0019-4197}
\email{xuelong_li@ieee.org}
\affiliation{%
  \institution{Institute of Artificial Intelligence, China Telecom (TeleAI)}
  \country{China}}
% \title{Unlocking Intrinsic Diversity in Autoregressive Video Distillation via Uncertainty Injection}

\title{Uncertainty DMD: Restoring Diversity in Few-Step Autoregressive Video Distillation}

\begin{abstract}
% Few-step distillation accelerates 
Few-step distillation improves the efficiency of autoregressive (AR) video generation, but often causes diversity collapse: under the same prompt, different noise samples tend to produce highly similar videos with weakened motion dynamics. We analyze this degradation in Distribution Matching Distillation
(DMD)-distilled AR video generators and find that, in the
autoregressive setting, it takes the form of a structured
\emph{uncertainty collapse}: the mode-seeking bias of DMD maps different noise samples
to nearly identical first chunks, and the deterministic AR cache then
propagates this collapsed state to all subsequent chunks, turning a
local loss of stochasticity at the rollout root into a global
suppression of temporal variation. Based on this analysis, we propose
\textbf{Uncertainty DMD}, a simple uncertainty-injection framework that
restores stochasticity at two key stages of AR generation: a timestep
perturbation for the first chunk to increase first-chunk diversity,
and a stochastic cache-writing mechanism for later chunks to preserve uncertainty in autoregressive conditioning. The method requires no architectural changes and introduces only lightweight perturbation operations. The same perturbation mechanisms are used during both training and inference. Experiments show that Uncertainty DMD consistently improves diversity and motion dynamics while maintaining comparable per-sample visual quality.
\noindent\textbf{Project page:} \url{https://scdzx.github.io/Uncertainty-DMD}

\end{abstract}


\begin{teaserfigure}
  \centering
    \vspace{-1.2em}
  \includegraphics[width=\textwidth,page=1]{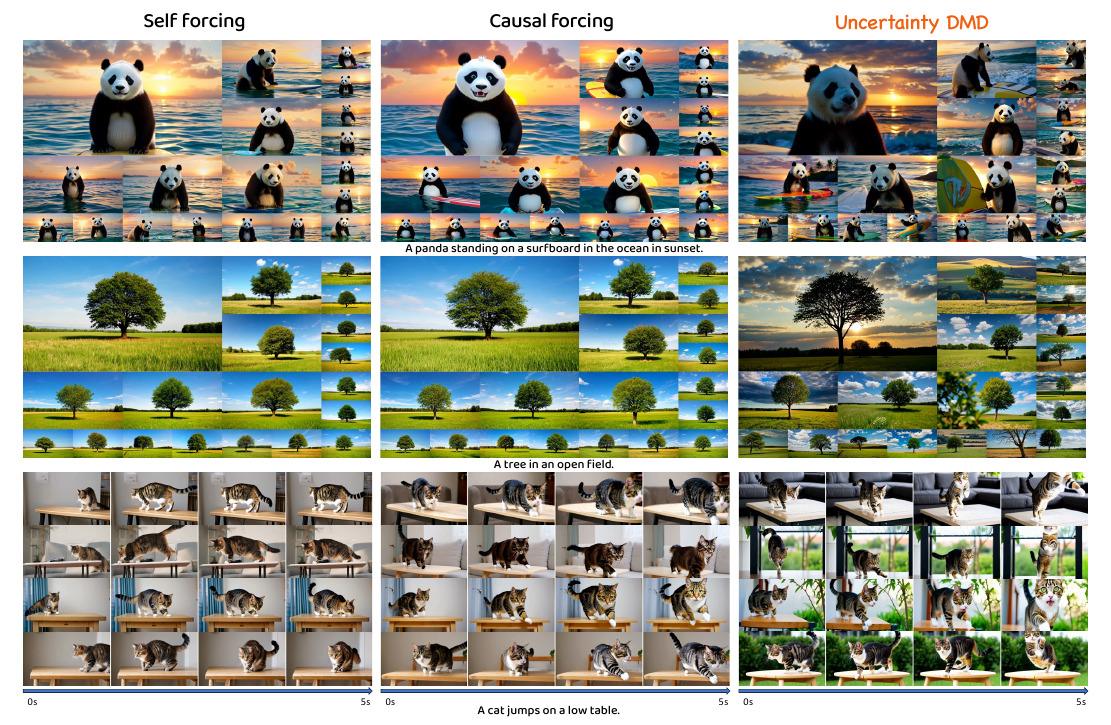}
  \vspace{-2.5em}
      \caption{\rev{Comparison using 19 random seeds ($0$--$18$) under the same setup. The first two rows show the first frame of each generated video for cross-seed comparison. In the third row, each horizontal strip represents one video, with frames ordered from left to right.}}
  \label{fig:teaser}
\end{teaserfigure}

\maketitle

\section{Introduction}

% Recent video diffusion models have achieved remarkable progress in generating realistic, temporally coherent, and text-aligned videos~\cite{hong2023cogvideo, jia2026moga, teng2025magi, hu2024animate}.
% However, scaling them to long videos remains challenging.
% Bidirectional video diffusion models denoise all frames jointly, which provides strong global consistency but incurs memory and compute costs that grow rapidly with video length~\cite{kong2024hunyuanvideo, wan2025}.
% Autoregressive video generation offers a more scalable alternative by producing videos chunk by chunk, where each new chunk is conditioned on a history cache of previously generated content~\cite{yin2025slow, huang2025self, helios}.
% This design amortizes the cost across chunks but does not eliminate the per-chunk overhead of multi-step diffusion sampling.
% Few-step distillation methods such as DMD~\cite{yin2024one} are therefore essential for turning powerful multi-step video generators into real-time autoregressive students.

% However, this acceleration comes at a cost.
% As shown in Fig.~\ref{fig:teaser}, after DMD, both Self-Forcing~\cite{huang2025self} and Causal Forcing~\cite{zhu2026causal} tend to produce visually similar videos across different random seeds under the same prompt.
% While each individual sample remains plausible, the diversity across samples drops sharply.
% That is, few-step distillation largely preserves per-sample quality but substantially reduces \emph{per-prompt sample diversity}---a key property of the original multi-step generator~\cite{wu2026diversity}.

\rev{Recent advances in AI have enabled more capable models~\cite{shao2025ai,an2026ai,lu2025law,huang2024domainfusion,wen2025metricsolverslidinganchoredmetric,Zhang_2026_CVPR,chen2026full4dgeneratingfullscope4d,gu2026searchtoworldevaluation3dworld}. Progress in image editing, audio-visual generation, and video compression has also advanced visual generative systems~\cite{zang2026instruction,song2026syncdit,chen2026generative,yuan2026enhancing,chen2024cascadezero123imagehighlyconsistent,chen2024liftimage3dliftingsingleimage,xiang2025makeefficientdynamicsparse,huang2025zero}.}
\rev{DMD is a distillation technique that transfers the generation capability of a multi-step diffusion teacher to a faster student requiring only a few denoising steps~\cite{yin2024one, yin2024improved, nguyen2024swiftbrush, shen2025efficient}. It has therefore been widely adopted for efficient image and video generation~\cite{lin2026autoregressive, chen2025sana, sauer2024adversarial, geng2026mean}.}
However, its distribution matching objective is mode-seeking: the student may concentrate on a few high-density modes rather than covering the full teacher distribution, resulting in reduced sample diversity~\cite{wu2026diversity, gandikota2026distilling, luo2024one}.
This issue is particularly relevant to autoregressive video generation, where long videos are generated chunk by chunk using a history cache~\cite{yin2025slow, huang2025self, helios}.
To reduce the diffusion sampling cost for each chunk, recent real-time autoregressive video generators often adopt DMD-based few-step distillation~\cite{huang2025self, zhu2026causal}, and therefore inherit its mode-seeking tendency.
Moreover, diversity collapse in early chunks can be stored in the cache and propagated to subsequent chunks, amplifying the loss of sample diversity throughout the video.

We observe this effect in DMD-distilled autoregressive video generators.
As shown in Fig.~\ref{fig:teaser}, after DMD, both Self-Forcing~\cite{huang2025self} and Causal Forcing~\cite{zhu2026causal} tend to produce visually similar videos across different random noise under the same prompt.
Although each individual video can still look plausible, the generated samples become much less diverse.
In other words, few-step DMD preserves per-sample quality to a large extent, but substantially reduces \emph{per-prompt sample diversity and motion dynamics}, which is an important property of the original multi-step generator.

\begin{figure}[t]
    \centering
    % 注意这里的 page=2 表示只加载该 PDF 的第二页
    \includegraphics[width=\linewidth, page=1]{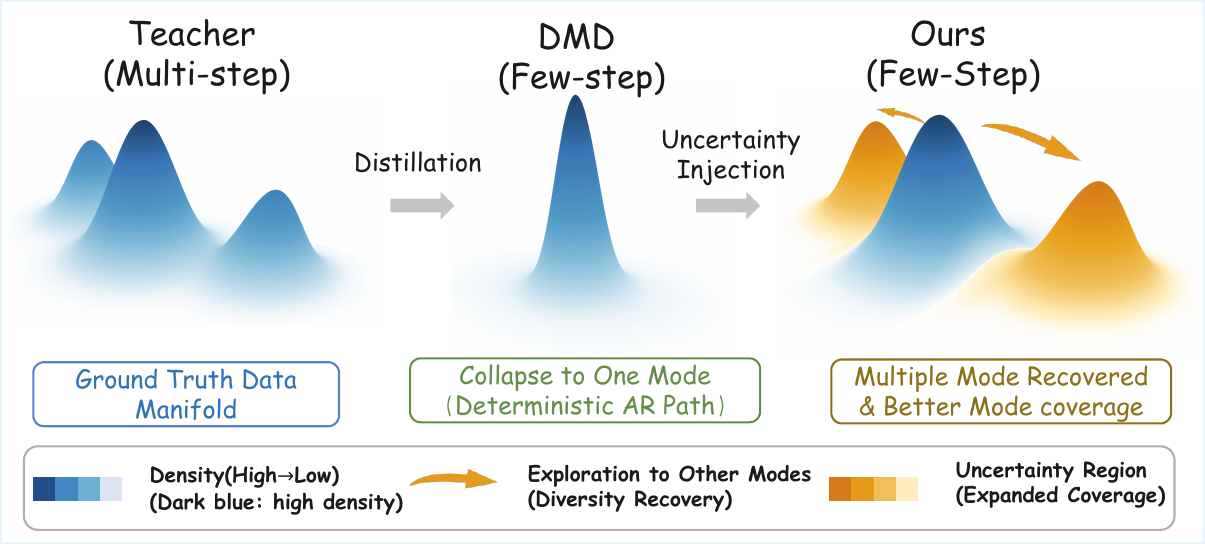} 
    \vspace{-1.5em}
    \caption{Conceptual illustration of our distributional reshaping. While standard DMD distillation collapses the multi-modal ground-truth manifold into a deterministic single mode (middle), our Uncertainty Injection strategy preserves and expands mode coverage (right).}
    \label{fig:mode_coverage}
    \vspace{-1.5em}
\end{figure}

This loss of diversity has both an \emph{algorithmic} and a \emph{structural} origin, and is not merely image-style mode collapse.
\textbf{Algorithmically}, DMD minimizes a reverse-KL-style objective
against the teacher, which is known to be \emph{mode-seeking}: the
student concentrates probability mass on a few high-density modes
rather than covering the full distribution~\cite{liu2026decoupled,
wu2026diversity}. In AR video generation, different initial noises yield similar scenes, and motion is suppressed across time. \textbf{Structurally}, the autoregressive rollout amplifies both effects: the first chunk is generated without prior context and written into
the history cache, which conditions all subsequent chunks and thereby fixes the scene layout, object placement, and coarse motion. Once different seeds collapse to similar first chunks, the resulting caches also collapse, while the per-step static bias further compresses temporal evolution.
As conceptually illustrated in Fig.~\ref{fig:mode_coverage}, this process progressively reduces mode coverage and drives the rollout toward a narrow deterministic trajectory.
We call this failure mode \textbf{uncertainty collapse}: mode-seeking distillation eliminates stochasticity both at the \emph{root} of the rollout (collapsing first chunks across noise) and along the \emph{temporal axis} (collapsing motion within each video), and the deterministic cache propagates both forms of collapse forward, locking the rollout into a narrow set of near-static trajectories.

% Motivated by this two-dimension diagnosis, we propose \textbf{Uncertainty DMD}, a structured uncertainty-injection framework for few-step autoregressive video distillation.
% Rather than perturbing raw latents in an unstructured manner, Uncertainty DMD reactivates two \emph{structured} stochasticity dimension that are squeezed by few-step DMD: the timestep condition at the root chunk and the autoregressive cache that carries history into future chunks.
% The former \emph{creates} seed-dependent branches at the root of the rollout, while the latter \emph{preserves} them as the rollout unfolds.
% Crucially, Uncertainty DMD are activated consistently across DMD training and inference, so that the student is explicitly optimized under the same stochasticity it will encounter at test time, avoiding the train--test mismatch of naive noise-injection. This design recovers per-prompt diversity at negligible overhead and requires no architectural modification.
Motivated by this two-dimensional diagnosis, we propose \textbf{Uncertainty DMD}, a structured uncertainty-injection framework for few-step autoregressive video distillation.
Instead of perturbing raw latents in an unstructured manner, Uncertainty DMD restores the lost uncertainty in few-step DMD by operating on two structured handles: the first-chunk timestep condition and the autoregressive cache that carries history into future chunks.
The former diversifies the first chunk by perturbing its denoising timestep, while the latter preserves the resulting diversity across subsequent chunks by injecting stochasticity into the autoregressive cache. Notably, the same uncertainty injection is applied during both DMD training and inference, so that the student is optimized under the stochasticity it encounters at test time, avoiding the train–test mismatch of naive noise injection.
This design recovers per-prompt diversity at negligible overhead and requires no architectural modification.

% In summary, our contributions are threefold:
% \begin{itemize}
%     \item \textbf{Problem identification and analysis.} We identify \emph{uncertainty collapse} as an intrinsic failure mode of DMD-distilled autoregressive video generators, manifesting as both reduced cross-seed diversity and suppressed temporal motion. We further pinpoint its origin as a two-level effect: mode-seeking distillation collapses the root chunk across seeds and biases each step toward static continuations, while the deterministic cache propagates both effects forward.

%     \item \textbf{A structured uncertainty-injection framework.} Guided by this diagnosis, we propose Uncertainty DMD, which reactivates two stochasticity dimensions squeezed by few-step DMD---the root-chunk timestep and the autoregressive cache---to create and preserve seed-dependent branches along the rollout, applied consistently across training and inference at negligible cost.

%     \item \textbf{Empirical validation.} We validate Uncertainty DMD on two representative DMD-based AR video generators, achieving consistent gains in diversity and motion dynamics while preserving per-sample visual quality and real-time efficiency. Compared with GAN-based distillation, naive noise injection, and vanilla DMD augmented with the same perturbations, our structured two- dimension design yields a more favorable and \emph{stable} diversity--quality trade-off throughout training, with ablations confirming that both dimensions are necessary and complementary.

% \end{itemize}
In summary, our contributions are threefold:
\begin{itemize}
    \item \textbf{Problem identification and analysis.}
    We identify \emph{uncertainty collapse} in DMD-distilled autoregressive video generators, where few-step distillation reduces diversity and suppresses motion. Through variance decomposition, cache transplantation, and teacher--DMD replacement studies, we show that uncertainty loss emerges in the first chunk and is further propagated by the autoregressive cache

    \item \textbf{Uncertainty-injection framework.}
    We propose Uncertainty DMD, which restores two structured sources of stochasticity suppressed by few-step DMD: the first-chunk timestep and the autoregressive cache. This creates and preserves diverse rollout branches with negligible overhead.

    \item \textbf{Empirical validation.}
    We validate Uncertainty DMD on two DMD-based AR video generators. Our analysis further reveals a trade-off in standard DMD distillation, where visual quality improves over training while diversity and motion dynamics progressively degrade. Uncertainty DMD substantially mitigates this degradation, improving diversity and motion dynamics while preserving visual quality, all without modifying the original training and generation framework.

\end{itemize}

\section{Related Work}
\noindent\textbf{Autoregressive Video Generation.} 
While recent video diffusion models achieve impressive visual quality, most rely on bidirectional or full-sequence denoising, which is costly and ill-suited for streaming or long-horizon synthesis~\cite{wan2025, kong2024hunyuanvideo, blattmann2023align, hong2023cogvideo, jia2026moga, teng2025magi, hu2024animate,zhang2026physomniphysicsgroundedmultiobjectscene,zhang2026tourphysicsbringingphysicsworld,wang2026directingworldfastautoregressive,huang2026cineweavertrainingfreereferencecontrollablemultishot}. Autoregressive video generation instead synthesizes frames or chunks sequentially conditioned on previously generated content, enabling low-latency streaming and scalable long video synthesis~\cite{zhu2026causal, xiang2025macro, chen2025teleworld, yin2025slow, helios, cui2026lol,xiang2026videoweaveunlockinggeometricconsistency}. To address the train--test mismatch and temporal drift inherent in causal generation, recent works explore self forcing training~\cite{huang2025self}, rolling windows, sink tokens, and explicit memory mechanisms~\cite{liu2026rolling, yang2026longlive, huang2025self, worldplay2025, xiang2026pathwise, yu2025context}, typically at the cost of architectural changes or additional retraining. Unlike these works that mainly target efficiency and temporal consistency, we focus on the intrinsic \emph{diversity collapse} of distilled autoregressive video generators.

\noindent\textbf{\rev{Few step Distillation and diversity collapse.}}
\rev{Consistency Models~\cite{song2023consistencymodels} preserve diversity better but achieve lower few-step quality.}
DMD accelerates diffusion models by matching the output distribution of a few-step student to that of a multi-step teacher~\cite{yin2024one, zhou2025adversarial, luo2023diff, fuest2026diffusion}, and improved variants such as DMD2~\cite{yin2024improved} achieve strong few-step generation quality. However, recent studies show that DMD-style objectives can suppress sample diversity due to their mode-seeking behavior. \rev{For image generation, f-DMD~\cite{xu2025one} introduces an additional discriminative score for \(f\)-divergence distillation, while ADM~\cite{lu2025adversarial} adopts GAN-based few-step distillation.} DP-DMD~\cite{wu2026diversity} mitigates this issue in image generation by preserving diversity at the first distilled step with a target-prediction objective, while applying the DMD loss only to later refinement steps. Diversity Distillation~\cite{gandikota2026distilling} observes that distilled models retain diversity-related representations but fail to activate them, and recovers diversity by skipping the first step or invoking an additional base model for the first denoising step, such first-step interventions may trade off sample quality. In video generation, DPP-GRPO~\cite{kazimi2025diverse} promotes video diversity by training a prompt-space policy to generate diverse prompt variants for a given input prompt, and evaluates diversity using metrics such as TCE~\cite{ibarrola2024measuring}, TIE~\cite{ibarrola2024measuring}, and VENDI~\cite{friedman2023the}. This prompt-level diversification is complementary to our fixed-prompt setting, where we restore diversity within the generation process itself. Different from these works, we study diversity collapse in DMD-based \emph{autoregressive} video generation, where collapsed early chunks can propagate through the autoregressive cache and constrain future temporal evolution.

\begin{figure}[t]
    \centering
    \includegraphics[width=\linewidth, page=1]{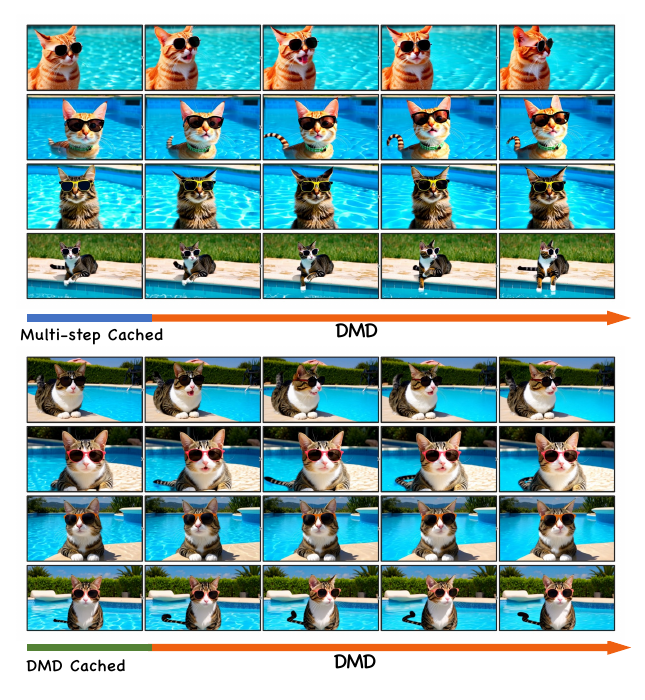}
      \vspace{-2.4em}
    \caption{
        \textbf{Cache Transplantation.}
        Four videos generated by the same DMD student with identical noise seeds for all subsequent chunks; only the first-chunk cache differs.
        \textbf{Top:} first-chunk cache from a multi-step teacher (\textit{Multi-step Cached}).
        \textbf{Bottom:} first-chunk cache from the DMD student itself (\textit{DMD Cached}).
        Prompt: \textit{``a cat wearing sunglasses in the pool''}.
    }
    \label{fig:cache_transplantation}
\end{figure}

\section{Diagnosing Mode Collapse in Few-Step AR Distillation}
\label{sec:prelim}

\subsection{Few-Step AR Video Generation and Diversity Evaluation}
\label{sec:prelim_ar}

AR video generation represents a video as a sequence of
temporal chunks $x_{1:L}=\{x_1,\ldots,x_L\}$ and generates them according to
the chunk-wise~\cite{yang2026longlive, huang2025self} conditional factorization
\begin{equation}
p_\theta(x_{1:L}\mid c)
=
p_\theta(x_1\mid c)
\prod_{i=2}^{L}
p_\theta(x_i\mid x_{<i},c),
\label{eq:ar_factor}
\end{equation}
where $c$ denotes the text condition.

In practical AR video models, previously generated chunks are stored in an
autoregressive cache $\mathcal{C}_{i-1}$. A $K$-step distilled student generates
the $i$-th chunk as
\begin{equation}
\hat{x}_{i,0}
=
G_\theta^{(K)}
\big(
z_i,\,
t_{1:K},\,
\mathcal{C}_{i-1},\,
c;\,
\epsilon_{1:K}
\big),
\label{eq:student_sampler}
\end{equation}
where $z_i$ is the initial noise, $t_{1:K}$ is the few-step timestep schedule,
and $\epsilon_{1:K}$ denotes the stochasticity during sampling. After each
chunk is generated, the cache is updated by a deterministic write operation:
\begin{equation}
\mathcal{C}_{i}
=
\operatorname{Append}
\big(
\mathcal{C}_{i-1},\hat{x}_{i,0}
\big).
\label{eq:cache_update}
\end{equation}

To evaluate generation diversity under the same prompt, we generate multiple
videos with different random seeds, extract video-level features, and compute
truncated entropy based on the covariance spectrum of these features~\cite{ibarrola2024measuring} and the spectrum of the sample-similarity kernel~\cite{friedman2023the}. Higher values indicate
that the generated videos are more dispersed in feature space and therefore
exhibit stronger diversity. Following DPP-GRPO~\cite{kazimi2025diverse}, we adopt the same computation protocol and extend it to six diversity metrics; detailed definitions and implementation details are provided in Supplementary.

\subsection{DMD Mode Collapse and First Chunk Collapse}
\label{sec:root_chunk_collapse}

Few-step students are often trained with DMD to match a multi-step teacher~\cite{yin2024one, yin2024improved}.
Given a generated sample $x=G_\theta(z,c)$ and its noisy version $x_t$, the DMD
gradient can be written as
% \begin{equation}
% \nabla_\theta \mathcal{L}_{\mathrm{DMD}}
% \approx
% \mathbb{E}_{z,t}
% \left[
% w(t)
% \big(
% s_{\mathrm{fake}}(x_t,t,c)
% -
% s_{\mathrm{real}}(x_t,t,c)
% \big)
% \frac{\partial G_\theta(z,c)}{\partial \theta}
% \right],
% \label{eq:dmd_grad}
% \end{equation}
\begin{equation}
\nabla_\theta \mathcal{L}_{\mathrm{DMD}}
=
\mathbb{E}_{z,t}
\left[
w(t)
\big(
s_{\mathrm{fake}}(x_t,t,c)
-
s_{\mathrm{real}}(x_t,t,c)
\big)
\frac{\partial G_\theta(z,c)}{\partial \theta}
\right],
\label{eq:dmd_grad}
\end{equation}
\rev{where $s_{\mathrm{real}}$ is the frozen teacher score and $s_{\mathrm{fake}}$
is estimated by an auxiliary fake-score model. Following~\cite{huang2025self},
$w(t)=1/\operatorname{mean}(|x-s_{\mathrm{real}}(x_t,t,c)|)$ normalizes the gradient magnitude.} Similar to the model-collapse
phenomenon observed in image-level DMD, this score-difference update tends to
encourage the student to concentrate on high-density teacher regions, which can
reduce sample diversity under different random noise\rev{~\cite{wu2026diversity, gandikota2026distilling, luo2024one}}.

This effect is particularly visible in the first chunk of AR video generation.
The first chunk has no historical cache condition,
\begin{equation}
\mathcal{C}_0 = \emptyset ,
\end{equation}
so global factors such as scene layout, object pose, camera viewpoint, and
initial motion must be determined only from the prompt and random noise:
\begin{equation}
\hat{x}_{1,0}
=
G_\theta^{(K)}
\big(
z_1,\,
t_{1:K},\,
\mathcal{C}_0,\,
c;\,
\epsilon_{1:K}
\big).
\label{eq:first_chunk_generation}
\end{equation}
% Under the mode-seeking bias of few-step DMD, different seeds may be mapped to
% similar high-probability first chunks,
\rev{Empirically, under the mode-seeking bias of few-step DMD, different seeds may be mapped to similar high-probability first chunks,}
\begin{equation}
G_\theta^{(K)}
\big(
z_1^a,t_{1:K},\mathcal{C}_0,c
\big)
\approx
G_\theta^{(K)}
\big(
z_1^b,t_{1:K},\mathcal{C}_0,c
\big),
\qquad
z_1^a \neq z_1^b .
\label{eq:root_chunk_collapse}
\end{equation}
We call this phenomenon \emph{first-chunk collapse}: for the same prompt,
different noise already produce similar layouts, poses, viewpoints, or initial
motion patterns in the first generated chunk.

\subsection{Deterministic Cache Propagates Diversity Collapse}
\label{sec:cache_propagation}

For later chunks, the student generates under the historical cache condition:
\begin{equation}
\hat{x}_{i,0}
=
G_\theta^{(K)}
\big(
z_i,\,
t_{1:K},\,
\mathcal{C}_{i-1},\,
c;\,
\epsilon_{1:K}
\big),
\qquad i>1 .
\label{eq:later_chunk_generation}
\end{equation}

\begin{figure}[t]
    \centering
    \includegraphics[width=0.8\linewidth, page=1]{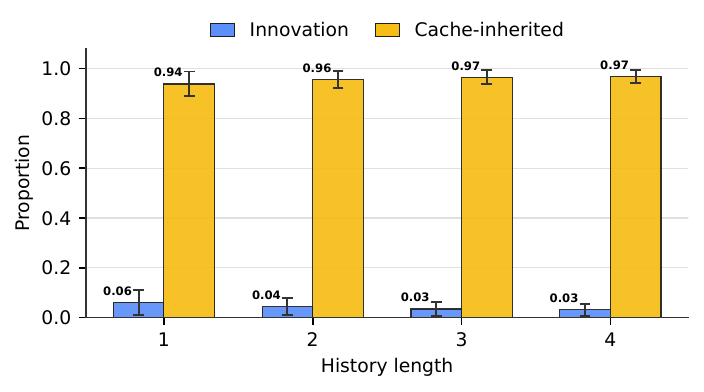}
      \vspace{-1em}
    \caption{
    Empirical variance decomposition of later-chunk diversity in DMD-based
    AR video generation. For each prompt and each history length, we
    sample 8 different history caches and 8 continuations for each cache.
    }
\label{fig:variance_decomposition}
\end{figure}

Thus, the diversity of a later chunk can be decomposed into two parts: the
variation produced by the current few-step sampler under a fixed cache, and
the variation caused by different historical caches. Let $\phi(\cdot)$ denote the feature of a generated chunk. By total variance
decomposition,
\begin{align}
\mathrm{Var}
\big[
\phi(\hat{x}_{i,0})
\mid c
\big]
&=
\mathbb{E}_{\mathcal{C}_{i-1}}
\left[
\mathrm{Var}
\big[
\phi(\hat{x}_{i,0})
\mid
\mathcal{C}_{i-1},c
\big]
\right]
\nonumber \\
&\quad+
\mathrm{Var}_{\mathcal{C}_{i-1}}
\left[
\mathbb{E}
\big[
\phi(\hat{x}_{i,0})
\mid
\mathcal{C}_{i-1},c
\big]
\right].
\label{eq:variance_decomposition}
\end{align}
The first term is the \emph{innovation diversity}, which measures how much new
semantic or motion variation the current sampler can still introduce when the
history cache is fixed. The second term is the \emph{cache-inherited diversity},
which measures how much different cache states lead to different future
continuations.

We further estimate the two terms in Eq.~\eqref{eq:variance_decomposition}
on DMD-based autoregressive video generation. We use 100 prompts and vary the
history chunk length $h \in \{1,2,3,4\}$, where each history chunk contains 3 latent
frames and each rollout is fixed to 7 chunks. For each prompt and each history
length, we first generate 8 different history caches
$\{\mathcal{C}^{a}_{h}\}_{a=1}^{8}$, and for each cache we generate 8 different
next-chunk continuations using different continuation noise. This gives
$8 \times 8 \times 4$ generated videos for each prompt. The variance under the
same cache estimates the innovation term, while the variance across different
cache means estimates the cache-inherited term.

As shown in Fig.~\ref{fig:variance_decomposition}, the cache-inherited term
dominates the total variance for all history lengths. It accounts for
$0.94$, $0.96$, $0.97$, and $0.97$ of the total variance when the history length
increases from 1 to 4, while the innovation term only contributes a small
fraction. This result indicates that, in few-step DMD autoregressive generation,
the historical cache is the main source controlling the diversity of later
chunks. Once different noise are mapped to similar cache states, the later
rollout has limited ability to recover diversity only through the current
sampler.

Few-step DMD first reduces first-chunk diversity, as the sampler becomes
less sensitive to random noise. Once similar first chunks are written into
the AR cache, later few-step chunks are conditioned on similar histories and
continue to inherit the collapsed state. Since the cache update is deterministic,
\begin{equation}
\mathcal{C}_{i}
=
\operatorname{Append}
\big(
\mathcal{C}_{i-1},
\hat{x}_{i,0}
\big),
\label{eq:deterministic_cache_update}
\end{equation}
the cache cannot create new high-level semantic branches by itself; it mainly
propagates the diversity, or lack of diversity, already present in previous
chunks.

Fig.~\ref{fig:cache_transplantation} illustrates this effect through cache
transplantation. We use the same DMD student and keep identical noise for
all subsequent chunks, changing only the first-chunk cache. When the cache comes
from the multi-step teacher, later chunks inherit more diverse historical
conditions. In contrast, when the cache comes from the DMD student itself, the
subsequent generations become much more similar. This shows that diversity
collapse in few-step AR distillation is not only a local sampling issue, but is
propagated along the rollout by the deterministic AR cache.

\begin{table}[t]
\centering
\small
\caption{Diagnosing diversity collapse on Causal Forcing.
Sampler choice for the first vs.\ later chunks isolates where diversity is lost.}
\label{tab:diagnosis}
\resizebox{0.8\linewidth}{!}{%
\begin{tabular}{l ccc ccc}
\toprule
& \multicolumn{3}{c}{Truncated Entropy $\uparrow$}
& \multicolumn{3}{c}{VENDI $\uparrow$} \\
\cmidrule(lr){2-4}\cmidrule(lr){5-7}
Configuration
& CLIP & INC & DINO
& CLIP & INC & DINO \\
\midrule
Teacher / Teacher   & 22.1 & 39.4 & 72.5 & 2.3 & 3.5 & 4.9 \\
Teacher / DMD  & 21.4 & 39.1 & 71.8 & 2.2 & 3.4 & 4.6 \\
DMD / Teacher  & 19.2 & 37.1 & 69.2 & 2.0 & 3.1 & 3.6 \\
DMD / DMD & 19.8 & 37.4 & 69.3 & 2.0 & 3.0 & 3.6 \\
\bottomrule
\end{tabular}%
}
\end{table}
\subsection{Testable Prediction and Motivation}
\label{sec:testable_prediction}

\begin{figure*}[t]
    \centering
    \includegraphics[width=1\textwidth]{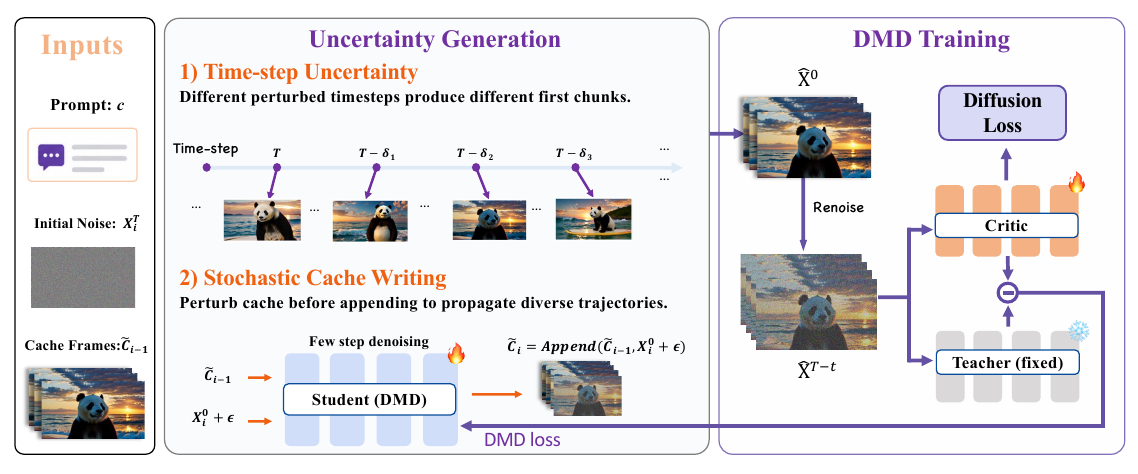}
    \vspace{-1.5em}
    \caption{\rev{
Overview of the proposed Uncertainty Injection DMD framework. 
Compared with standard DMD, our method perturbs the structured autoregressive state by injecting timestep uncertainty into the first chunk and stochastic noise into cache writing.
}}
    \label{fig:quality_diversity_training}
\end{figure*}

The above analysis gives a direct testable prediction. If diversity collapse
mainly originates from the first chunk and is then propagated by the AR cache,
then replacing only the first-chunk sampler should recover most of the lost
diversity. In contrast, replacing only the later-chunk sampler should provide
limited improvement, because later chunks are still conditioned on a collapsed
first cache.

Let $T$ denote the multi-step teacher sampler and $Q$ denote the few-step DMD
student. We expect
\begin{equation}
\mathrm{Div}(T_1,Q_{>1})
-
\mathrm{Div}(Q_{1:L})
\gg
\mathrm{Div}(Q_1,T_{>1})
-
\mathrm{Div}(Q_{1:L}),
\label{eq:first_later_prediction}
\end{equation}
where $\mathrm{Div}(\cdot)$ denotes a cross-sample diversity metric, such as
Truncated Entropy~\cite{ibarrola2024measuring} or VENDI~\cite{friedman2023the}. Here, $(T_1,Q_{>1})$ means using the teacher only
for the first chunk and the DMD student for later chunks, while $(Q_1,T_{>1})$
means using the DMD student for the first chunk and the teacher for later
chunks.

Table~\ref{tab:diagnosis} validates this prediction. Replacing only the first
chunk with the teacher sampler (\textit{Teacher / DMD}) recovers most of the
diversity compared with the full DMD rollout (\textit{DMD / DMD}). In contrast,
replacing only later chunks (\textit{DMD / Teacher}) brings much smaller gains,
showing that the collapsed first-chunk cache already constrains the subsequent
rollout.

% Taken together, the three analyses above point to a consistent picture.
% The first-chunk collapse experiment (Sec.~\ref{sec:root_chunk_collapse}) shows that few-step DMD
% maps different seeds to nearly identical first chunks; the variance
% decomposition (Fig.~\ref{fig:variance_decomposition}) shows that, once the cache is fixed, the
% sampler can no longer introduce meaningful innovation diversity
% (>94\% of later-chunk variance is cache-inherited); and the cache
% transplantation together with sampler-swap results (Fig.~\ref{fig:cache_transplantation}, Table~\ref{tab:diagnosis})
% confirm that the first chunk and the AR cache jointly act as the
% bottleneck of the entire rollout.

% While the reverse-KL-like, mode-seeking tendency of DMD has been well observed in image distillation~\cite{wu2026diversity, gandikota2026distilling}, AR video generation adds a structural amplification mechanism: the collapsed first chunk is written into the deterministic history cache and then conditions all subsequent chunks. Consequently, once the cache is fixed, perturbing intermediate latents in later denoising steps has limited ability to recover missing modes. Uncertainty should instead be injected into the two AR state variables that steer the rollout: the first-chunk denoising trajectory and the autoregressive cache writing. This directly motivates our Uncertainty DMD.
Taken together, the three analyses above reveal that diversity collapse in few-step AR distillation is not only the known reverse-KL-like, mode-seeking behavior of DMD observed in image distillation~\cite{wu2026diversity, gandikota2026distilling}, but is further amplified by the autoregressive structure. The first-chunk collapse experiment (Sec.~\ref{sec:root_chunk_collapse}) shows that few-step DMD maps different noise to nearly identical first chunks; the variance decomposition (Fig.~\ref{fig:variance_decomposition}) shows that, once the cache is fixed, the sampler contributes little innovation diversity, with more than 94\% of later-chunk variance inherited from the cache; and the cache transplantation and sampler-swap studies (Fig.~\ref{fig:cache_transplantation}, Table~\ref{tab:diagnosis}) confirm that the collapsed first chunk is written into the deterministic history cache and then constrains all subsequent chunks. Therefore, perturbing intermediate latents in later denoising steps has limited ability to recover missing modes. Instead, uncertainty should be injected into the two AR state variables that actually steer the rollout: the first-chunk denoising trajectory and the autoregressive cache writing. This directly motivates our Uncertainty DMD.

\section{Methodology}
\label{sec:method}

% We inject uncertainty into two autoregressive state variables: the timestep condition of the first chunk and the cache state written for later
% chunks. Both perturbations are used during training and inference; during
% training, their strengths are gradually increased by a curriculum schedule,
% while at inference we use the final scheduled probabilities. We describe timestep perturbation in Sec.~\ref{sec:module_a_root_timestep}, stochastic cache
% writing in Sec.~\ref{sec:module_b_cache_writing}, and the curriculum schedule
% with the overall algorithm in Sec.~\ref{sec:uncertainty_curriculum_algorithm}.
Uncertainty DMD injects stochasticity into the two AR state variables: the first-chunk timestep condition and the cache written for future chunks. Sec.~\ref{sec:module_a_root_timestep} introduces timestep perturbation, Sec.~\ref{sec:module_b_cache_writing} introduces stochastic cache writing, and Sec.~\ref{sec:uncertainty_curriculum_algorithm} gives the curriculum schedule and overall algorithm. The overall framework of Uncertainty DMD is illustrated in Fig.~\ref{fig:quality_diversity_training}.

\subsection{Module A: Timestep Perturbation}
\label{sec:module_a_root_timestep}

As described in Sec.~\ref{sec:prelim_ar}, the student generates
each video chunk with a few-step denoising schedule
\(\tau_1=T>\tau_2>\cdots>\tau_K\). In few-step distilled samplers, the first
denoising call largely fixes the coarse layout and motion, while later calls
mainly refine it. Because this call starts from almost pure noise, a fixed
timestep makes every sample take the same large denoising update, which can wash
out noise differences and push outputs toward similar modes. We perturb
this timestep condition so the first update is made under slightly different
noise levels, allowing different noise to choose different early modes. This also explains why skipping the first step in DMD image models~\cite{gandikota2026distilling} can
increase diversity: \rev{feeding a noisy latent with a randomly sampled cleaner timestep $T-\Delta t$ causes part of the unresolved noise to be interpreted as meaningful structural variation, leading to more diverse trajectories, while skipping the first step similarly avoids the mode-collapsing initial update.}

% In few-step distilled samplers, the first
% denoising call largely fixes the coarse layout and motion, while later calls
% mainly refine it. Because this call starts from almost pure noise, a fixed
% timestep makes every sample take the same large denoising update, which can wash
% out noise differences and push outputs toward similar modes. We perturb
% this timestep condition so the first update is made under slightly different
% noise levels, allowing different noise to choose different early modes. This also explains why skipping the first step in DMD image models~\cite{gandikota2026distilling} can
% increase diversity: it reduces the effect of the mode-collapsing initial update.

 The first chunk starts from the initial noisy latent
\(x_{1,T}\). When first timestep perturbation is activated, we keep this latent
unchanged but replace the timestep condition \(T\) by a perturbed timestep
\[
    \tilde{T}=T-\Delta,
    \qquad
    \Delta \sim p_{\Delta},
\]
\rev{where $\tilde{T}$ is clipped to the valid timestep range, and $p_{\Delta}$ is the
timestep-offset distribution, implemented as $\Delta \sim \mathcal{U}(0,100)$.
The first generator call for the first chunk becomes}
\[
    \hat{x}_{1,0}^{(1)}
    =
    G_\theta(x_{1,T}, \tilde{T}, \mathcal{C}_0, c),
    \qquad
    \mathcal{C}_0=\varnothing .
\]
Equivalently, the generator receives the same latent as in the vanilla
few-step sampler, but is conditioned on a slightly shifted noise level
\(T-\Delta\). This timestep mismatch changes the first-chunk denoising
trajectory without directly perturbing the latent itself, thereby inducing
different branches for the subsequent autoregressive rollout.

During training, this operation is applied to the generator rollout with
probability \(p_{\rm root}(s)\), which is gradually increased by the curriculum
schedule in Sec.~\ref{sec:uncertainty_curriculum_algorithm}. 

\subsection{Module B: Stochastic Cache Writing}
\label{sec:module_b_cache_writing}

For later chunks, we inject uncertainty into the autoregressive
conditioning through a stochastic cache-writing mechanism. Under
self-forcing training, the AR cache is constructed from the model's
own rollout. At autoregressive step $j$, the rollout reaches a sampled
truncation step and the model outputs a clean prediction
$\hat{x}_{j,0}$ for the current chunk. Before writing this predicted
clean chunk into the cache, we optionally perturb it once. After
insertion, the cached representation is kept fixed and is not
re-perturbed at later autoregressive steps.

We define the forward noising operator as
\begin{equation}
    \Psi(x,\epsilon,\tau)
    =
    \alpha_{\tau} x
    +
    \sigma_{\tau} \epsilon ,
\end{equation}
where $\alpha_{\tau}$ and $\sigma_{\tau}$ are predefined coefficients
associated with noise level $\tau$.

Before cache insertion, the predicted clean chunk is perturbed as
\begin{equation}
    \bar{x}_{j,0}
    =
    (1-m_j)\hat{x}_{j,0}
    +
    m_j \Psi(\hat{x}_{j,0},\epsilon_j,\tau_{\mathrm{cache}}),
\end{equation}
where
\begin{equation}
    m_j \sim \mathrm{Bernoulli}(p_{\mathrm{cache}}),
    \qquad
    \epsilon_j \sim \mathcal{N}(0,\mathbf{I}).
\end{equation}
Here, $p_{\mathrm{cache}}$ denotes the probability of perturbing the
predicted clean chunk before cache insertion, and
$\tau_{\mathrm{cache}}$ controls the perturbation strength through
$\alpha_{\tau_{\mathrm{cache}}}$ and
$\sigma_{\tau_{\mathrm{cache}}}$.

The perturbed chunk is then written into the AR cache:
\begin{equation}
    \tilde{\mathcal{C}}_{j}
    =
    \mathrm{Append}
    \left(
        \tilde{\mathcal{C}}_{j-1},
        \bar{x}_{j,0}
    \right),
    \qquad
    \tilde{\mathcal{C}}_{0}=\varnothing .
\end{equation}
Thus, when generating the $i$-th chunk, the cache
$\tilde{\mathcal{C}}_{i-1}$ contains fixed entries
$\{\bar{x}_{1,0},\bar{x}_{2,0},\dots,\bar{x}_{i-1,0}\}$, each sampled
and written only once during the rollout. The student predicts the
current chunk conditioned on this perturbed AR cache:
\begin{equation}
    \hat{x}_{i,0}
    =
    G_\theta(x_{i,t},t,\tilde{\mathcal{C}}_{i-1},c),
    \qquad i>1 .
\end{equation}

The same stochastic cache-writing rule is used in both training and
inference. During training, the cache is recursively constructed from
previously generated clean predictions along the self-forcing rollout.
During inference, each newly generated chunk is optionally perturbed
once before cache insertion and then kept fixed for all subsequent
autoregressive steps.

\subsection{Uncertainty Curriculum and Overall Algorithm}
\label{sec:uncertainty_curriculum_algorithm}
We apply the two structured perturbations with a simple curriculum schedule.
At the early stage of DMD training, both perturbations are disabled. After a
warm-up iteration $S_w$, their probabilities are gradually increased with the
training iteration. Let $\ell$ denote the current training iteration. We define
\[
\gamma_\ell =
\mathrm{clip}\left(\frac{\ell-S_w}{R},0,1\right),
\]
where $R$ is the ramp-up length. The probabilities of root timestep
perturbation and stochastic cache writing are
\[
p_{\rm root}(\ell)=\gamma_\ell,
\qquad
p_{\rm cache}(\ell)=\gamma_\ell p_{\rm cache}^{\max}.
\]
Thus, after the curriculum finishes, timestep perturbation is always
enabled, while cache writing perturbation is applied with a high probability
$p_{\rm cache}^{\max}$. At inference, we directly use the final probabilities,
i.e., $p_{\rm root}=1$ and $p_{\rm cache}=p_{\rm cache}^{\max}$. The overall
training procedure with the proposed structured uncertainty is summarized in
Alg.~\ref{alg:dmd_su}.

\definecolor{keyblue}{HTML}{2563EB}
\definecolor{keyred}{HTML}{DC2626}
\definecolor{keypurple}{HTML}{7C3AED}
\definecolor{keygreen}{HTML}{16A34A}

\section{Experiments}
\noindent\textbf{Implementation.} We implement our approach on two AR video generation frameworks, Causal Forcing~\cite{zhu2026causal} and Self-Forcing~\cite{huang2025self}. Both are built upon the Wan2.1-T2V-1.3B~\cite{wan2025} to generate 5-second videos at 16 FPS with a resolution of $832 \times 480$. Following baseline protocols, we adopt a chunk-wise AR scheme generating a chunk of 3 latent frames at a time~\cite{yin2025slow}, and utilize text prompts from a VidProM~\cite{wang2024vidprom} dataset for data-free DMD. All models are trained on 32 H100 GPUs with a batch size of 32. Unless otherwise specified, the default training duration is set to 600 steps. Additional experimental details, including dataset construction, training hyperparameters, inference settings, and metric computation, are provided in \rev{Supplementary}.

\begin{algorithm}[H]
\caption{DMD Training with Structured Uncertainty}
\label{alg:dmd_su}
\begin{algorithmic}[1]
\REQUIRE AR generator $G_\theta$, KV encoder $G_\theta^{\mathrm{KV}}$,
warm-up $S_w$, ramp length $R$, maximum cache probability
$p_{\rm cache}^{\max}$.

\FOR{training iteration $\ell=1,2,\dots$}
    \STATE Compute curriculum strength
    ${\color{keyblue}
    \gamma_\ell=\mathrm{clip}((\ell-S_w)/R,0,1)}$.
    \STATE Set
    ${\color{keyblue}
    p_{\rm root}=\gamma_\ell}$ and
    ${\color{keyblue}
    p_{\rm cache}=\gamma_\ell p_{\rm cache}^{\max}}$.
    \STATE Sample training chunks and initialize AR cache
    $\mathrm{KV}\leftarrow []$.

    \FOR{chunk index $i=1,\dots,K$}
        \STATE Sample $u\sim\mathcal{U}(0,1)$.
        \IF{$i=1$ and $u<p_{\rm root}$}
            \STATE Replace the first denoising timestep
            ${\color{keyred}
            \tau_1=T \rightarrow
            \tilde{\tau}_1=T-\Delta,\ \Delta\sim p_\Delta}$.
            \hfill \COMMENT{\textcolor{keyred}{root timestep perturbation}}
        \ENDIF

        \STATE Generate chunk $\hat{x}^i_0$ by the standard DMD rollout with
        the current schedule and cache $\mathrm{KV}$.

        \STATE Set $\bar{x}^i_0\leftarrow \hat{x}^i_0$.
        \STATE Sample $v\sim\mathcal{U}(0,1)$.
        \IF{$v<p_{\rm cache}$}
            \STATE
            ${\color{keypurple}
            \bar{x}^i_0
            \leftarrow
            \Psi(\hat{x}^i_0,\epsilon_i,\tau_{\rm cache})}$,
            ${\color{keypurple}
            \epsilon_i\sim\mathcal{N}(0,\mathbf{I})}$.
            \hfill \COMMENT{\textcolor{keypurple}{stochastic cache writing}}
        \ENDIF

        \STATE Write cache:
        $\mathrm{KV}
        \leftarrow
        \mathrm{KV}\cup
        G_\theta^{\mathrm{KV}}(\bar{x}^i_0,0,\mathrm{KV})$.
    \ENDFOR

    \STATE Update $G_\theta$ with the standard DMD objective.
\ENDFOR

\STATE \textcolor{keygreen}{\textbf{Inference:}}
use
${\color{keygreen}
p_{\rm root}=1}$ and
${\color{keygreen}
p_{\rm cache}=p_{\rm cache}^{\max}}$
with the same perturbation rules.

\end{algorithmic}
\end{algorithm}

\begin{table*}[t]
\centering
\caption{Quantitative comparison of video diversity and generation quality on the 2,000 generated videos}

\label{tab:main_results}
\vspace{-1.2em}
\resizebox{\textwidth}{!}{

\begin{tabular}{l cccccc c cccccc}
\toprule
\multirow{2}{*}{\textbf{Method}} 
& \multicolumn{6}{c}{\textbf{Video Diversity ($\uparrow$)}} 
& 
& \multicolumn{6}{c}{\textbf{Generation Quality (VBench) ($\uparrow$)}} \\
\cmidrule{2-7} \cmidrule{9-14}
& TE-CLIP & TE-Inc. & TE-DINO 
& VENDI-CLIP & VENDI-Inc. & VENDI-DINO
& 
& Sub. Cons. & Bg. Cons. & Motion Sm. & Dyn. Deg. & Aesthetic & Imaging \\
\midrule

\multicolumn{14}{l}{\textbf{Causal Forcing}} \\

DMD                 
 & 19.8 & 37.4 & 69.3 & 2.0 & 3.0 & 3.6
& 
& \textbf{98.1} & \textbf{97.4} & \textbf{99.1} & 31.1 & \textbf{67.4} & 73.3 \\

DMD + \textbf{Ours} 
& \textbf{22.1} & \textbf{39.4} & \textbf{71.9}
& \textbf{2.3} & \textbf{3.5} & \textbf{4.7}
& 
& 97.3 & 97.0 & 99.0 & \textbf{41.8} & 66.6 & \textbf{73.5} \\

\midrule
\multicolumn{14}{l}{\textbf{Self Forcing}} \\

DMD                   
& 16.4 & 33.5 & 66.1
& 1.6 & 2.4 & 2.6
& 
& \textbf{98.6} & 97.6 & 99.2 & 25.1 & 65.9 & 72.5 \\

DMD + GAN             
& 16.8 & 33.7 & 66.3
& 1.7 & 2.4 & 2.6
& 
& 97.8 & 96.8 & 98.9 & 40.4 &65.9 & 72.7 \\

DMD + Extra Noise             
& 16.1 &33.2  & 65.6
& 1.7 & 2.4 & 2.5
& 
& 97.9 & \textbf{97.9} & \textbf{99.3} & 15.9 & \textbf{66.5} & \textbf{73.2} \\

DMD + \textbf{Ours}   
& \textbf{19.5} & \textbf{36.1} & \textbf{68.6}
& \textbf{1.9} & \textbf{2.8} & \textbf{3.3}
& 
& 97.6 & 96.6 & 98.7 & \textbf{42.4} & 64.6 & \textbf{72.8} \\

\midrule

\textcolor{gray}{\textit{Wan (Bi-directional)}} 
& \textcolor{gray}{\textit{20.9}} 
& \textcolor{gray}{\textit{38.0}} 
& \textcolor{gray}{\textit{71.8}} 
& \textcolor{gray}{\textit{2.0}} 
& \textcolor{gray}{\textit{3.3}} 
& \textcolor{gray}{\textit{4.5}} 
& 
& \textcolor{gray}{\textit{98.2}} 
& \textcolor{gray}{\textit{98.4}} 
& \textcolor{gray}{\textit{99.1}} 
& \textcolor{gray}{\textit{22.7}} 
& \textcolor{gray}{\textit{65.7}} 
& \textcolor{gray}{\textit{71.3}} \\

\bottomrule
\end{tabular}
}

\raggedright

\end{table*}

\noindent\textbf{Evaluation.}
We use VBench~\cite{huang2024vbench} as our primary benchmark to evaluate both intrinsic diversity and generation quality. We randomly sample 100 text prompts from the VBench suite and generate 20 videos per prompt with a fixed set of random seeds shared across all methods, resulting in 2,000 videos for each method. Following DPP-GRPO~\cite{kazimi2025diverse}, we measure set-level diversity among videos generated from the same prompt. We extend the three diversity metrics in DPP-GRPO to six metrics by computing both Truncated Entropy~\cite{ibarrola2024measuring} and VENDI scores~\cite{friedman2023the} on three feature spaces: CLIP~\cite{radford2021learning}, Inception~\cite{szegedy2016rethinking}, and DINO~\cite{oquab2024dinov}. These metrics are denoted as TE-CLIP, TE-INCEPTION, TE-DINO, VENDI-CLIP, VENDI-INCEPTION, and VENDI-DINO. We further report the standard VBench quality metrics over all 2,000 generated videos.

\noindent\textbf{Quantitative results.}
Table~\ref{tab:main_results} reports a quantitative comparison of video diversity and generation quality. Our method consistently improves video diversity while maintaining comparable VBench quality. This shows that the proposed structured uncertainty effectively recovers noise-dependent variation in autoregressive video distillation. \rev{Restored diversity recovers more motion modes, allowing more
samples to pass the dynamic threshold underlying Dynamic Degree, a high-variance
binary metric; DMD + GAN shows a similar trend. Moreover, autoregressive models
can achieve higher Dynamic Degree than bidirectional baselines, as also observed
in Causal Forcing.} We also compare with two auxiliary baselines under the Self-Forcing
initialization. DMD + GAN only brings limited diversity improvement, and
DMD + Extra Noise does not provide consistent gains. Here, DMD + Extra Noise
directly adds Gaussian noise to the generator latent during training:
\[
    \tilde{x}_{i,t}=x_{i,t}+m_i \cdot 0.1\epsilon_i,
    \quad 
    m_i\sim\mathrm{Bernoulli}(0.1),\ 
    \epsilon_i\sim\mathcal{N}(0,I).
\]
In contrast, our method perturbs structured AR variables, namely the timestep and the historical cache, leading to more effective diversity recovery.
In Supplementary, we further show that our method outperforms two alternatives, i.e., skipping the first denoising step in the first chunk and replacing it with the multi-step teacher sampler, yielding better diversity recovery with comparable quality.We also conduct a user study, where our method receives comparable perceptual quality scores to standard DMD while obtaining substantially higher diversity scores.

\noindent\textbf{Ablation study.}
Table~\ref{tab:ablation} ablates the key components of our method under the Self-Forcing initialization. 
Removing timestep perturbation significantly reduces diversity, e.g., TE-CLIP drops from 19.5 to 16.7 and VENDI-DINO drops from 3.3 to 2.6, showing that injecting uncertainty into the initial chunk is crucial for preserving seed-dependent variations. 
Removing stochastic cache writing keeps most diversity metrics close to the full model, but lowers the dynamic degree from 42.4 to 37.0, indicating that perturbing the historical cache mainly helps maintain stronger temporal dynamics. 
Without the curriculum schedule, the model still improves over vanilla DMD, but both diversity and motion dynamics degrade compared with the full method: TE-CLIP decreases from 19.5 to 18.4, VENDI-DINO decreases from 3.3 to 3.1, and the dynamic degree drops from 42.4 to 37.4. 
This suggests that gradually introducing the proposed uncertainty perturbations stabilizes training and helps the model better adapt to stochastic initial states and noisy historical caches. 
Overall, the full method achieves the best balance between diversity and dynamic generation quality.

\begin{table}[t]
\centering

\caption{Ablation study of the proposed components.}
\vspace{-1em}

\label{tab:ablation}
\resizebox{\linewidth}{!}{
\begin{tabular}{lcccccc}
\toprule
\multicolumn{7}{l}{\textbf{(a) Video Diversity ($\uparrow$)}} \\
\midrule
\textbf{Method} 
& VENDI-CLIP & VENDI-Inc. & VENDI-DINO 
& TE-CLIP & TE-Inc. & TE-DINO \\
\midrule
Ours 
& 1.9 & 2.8 & 3.3 
& 19.5 & 36.1 & 68.6 \\

w/o Timestep Perturbation
& 1.7 & 2.5 & 2.6
& 16.7 & 33.3 & 65.7 \\

w/o Stochastic Cache Writing
& 1.8 & 2.8 & 3.2
& 19.1 & 36.0 & 68.5 \\

w/o Curriculum
& 1.8 & 2.8 & 3.1
& 18.4 & 35.5 & 68.0 \\

% DMD 
% & 1.6 & 2.4 & 2.6
% & 16.8 & 34.9 & 66.9  \\

\midrule
\multicolumn{7}{l}{\textbf{(b) Generation Quality (VBench) ($\uparrow$)}} \\
\midrule
\textbf{Method} 
& Sub. Cons. & Bg. Cons. & Motion Sm. 
& Dyn. Deg. & Aesthetic & Imaging \\
\midrule

Ours 
& 97.6 & 96.6 & 98.7 
& 42.4 & 64.6 & 72.8 \\

w/o Timestep Perturbation
& 98.3 & 97.3 & 98.9
& 32.6 & 66.3 & 72.7 \\

w/o Stochastic Cache Writing
& 98.4 & 97.3 & 98.9
& 37.0 & 65.3 & 73.0 \\

w/o Curriculum
& 97.6 & 96.8 & 98.8
& 37.4 & 64.9 & 72.5 \\

% DMD 
% & 98.6 & 97.6 & 99.2
% & 25.1 & 65.9 & 72.5 \\

\bottomrule
\end{tabular}
}
\vspace{-1.5em}
\end{table}

\noindent\rev{\textbf{Generalization study.} Following Causal Forcing~\cite{zhu2026causal}, we extend our method to DMD with the Wan2.1 1.3B frame-wise and 14B chunk-wise models. As shown in Table~\ref{tab:generalization-backbone}, the results show its generalizability across different model sizes and temporal granularities.}

\begin{figure*}
    \centering
    \includegraphics[width=\textwidth]{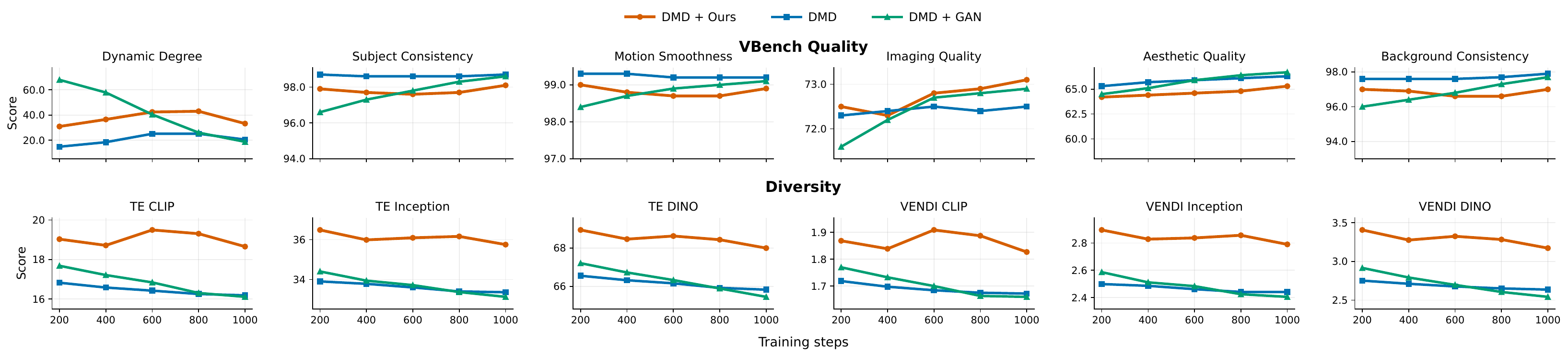}
    \caption{Quality and diversity metrics across training steps.}
    \label{fig:quality_diversity_training}
\end{figure*}

\begin{table}[t]
\centering
\revcolor

\caption{Comparison of diversity and quality under different model settings.}
\vspace{-1em}
\label{tab:generalization-backbone}
\resizebox{\linewidth}{!}{
\begin{tabular}{lcccccc}
\toprule
\multicolumn{7}{l}{\textbf{(a) Video Diversity ($\uparrow$)}} \\
\midrule
\textbf{Method}
& \textbf{VENDI-CLIP}
& \textbf{VENDI-Inc.}
& \textbf{VENDI-DINO}
& \textbf{TE-CLIP}
& \textbf{TE-Inc.}
& \textbf{TE-DINO} \\
\midrule
Frame-wise 1.3B + DMD 
& 1.9 & 2.8 & 3.2
& 18.4 & 35.4 & 67.8 \\

Frame-wise 1.3B + \textbf{Ours}
& 2.2 & 3.3 & 4.2
& 21.5 & 38.2 & 71.0 \\

Chunk-wise 14B + DMD 
& 1.8 & 2.7 & 3.0
& 17.3 & 34.9 & 67.2 \\

Chunk-wise 14B + \textbf{Ours}
& 2.2 & 3.4 & 4.4
& 21.8 & 39.0 & 71.5 \\

\midrule
\multicolumn{7}{l}{\textbf{(b) Generation Quality (VBench) ($\uparrow$)}} \\
\midrule
\textbf{Method}
& \textbf{Sub. Cons.}
& \textbf{Bg. Cons.}
& \textbf{Motion Sm.}
& \textbf{Dyn. Deg.}
& \textbf{Aesthetic}
& \textbf{Imaging} \\
\midrule
Frame-wise 1.3B + DMD 
& 98.3 & 97.0 & 99.3
& 21.1 & 65.7 & 72.6 \\

Frame-wise 1.3B + \textbf{Ours}
& 97.0 & 96.1 & 99.1
& 38.3 & 64.6 & 72.3 \\

Chunk-wise 14B + DMD 
& 98.7 & 98.1 & 99.4
& 32.9 & 66.6 & 72.3 \\

Chunk-wise 14B + \textbf{Ours}
& 98.3 & 97.1 & 99.1
& 36.6 & 64.6 & 72.5 \\

\bottomrule
\end{tabular}

}

\end{table}

\noindent\textbf{Training dynamics.} 
\rev{Figure~\ref{fig:quality_diversity_training} compares DMD, DMD + GAN, and
DMD + Ours at different training steps. All methods show similar quality trends
during training, while our method maintains competitive VBench quality and a
higher dynamic degree in later stages. For diversity, DMD + Ours consistently
outperforms DMD and DMD + GAN across training steps, indicating that the proposed
uncertainty injection slows down diversity degradation during distillation.}

\begin{figure}
    \centering
    % 注意这里的 page=2 表示只加载该 PDF 的第二页
    \includegraphics[width=\linewidth, page=1]{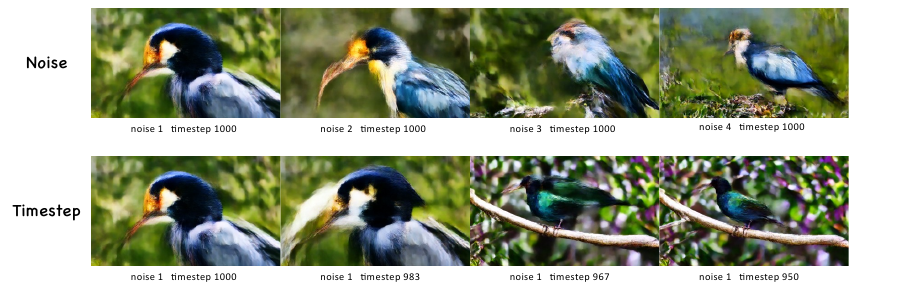} 
    \vspace{-1.5em}
    \caption{\rev{Visualization of noise and timestep effects on the first denoising step of the first frame using four-step Self-Forcing model. All results use the same prompt, \textit{``a beautiful bird.''} \textit{Noise}: different random noise samples at timestep 1000. \textit{Timestep}: four evenly spaced timesteps from 1000 to 950 (left to right), using the same noise sample.}}
    \label{fig:mode_coverage2}
\vspace{-1.5em}
\end{figure}
\noindent\textbf{Timestep perturbation.}
\rev{Figure~\ref{fig:mode_coverage2} shows the first-step denoising behavior of the few-step model with fixed initial noise and text prompt, while varying only the input timestep. The predictions vary noticeably, indicating that timestep perturbation introduces additional stochasticity and enables diverse generation trajectories from the same noise.}

\section{Conclusion}
\rev{We propose uncertainty DMD to mitigate diversity collapse in DMD-based autoregressive video diffusion distillation. By combining timestep perturbation with stochastic cache writing, our method improves sample diversity and motion dynamics while maintaining comparable visual quality. Nevertheless, current autoregressive video generation lacks explicit planning for future motion, which may result in unnatural complex motions.}

 \begin{acks} 
This work was supported in part by the Natinal Natural Science Foundation of China, under Grant Numbers 62192783, 62276128, and 62406140; the Young Elite Scientists Sponsorship Program by China Association for Science and Technology, under  Grant Number 2023QNRC001; the Key Research and Development Program of Jiangsu Province, under Grant Number BE2023019; and the Jiangsu Natural Science Foundation, under Grant Nos. BK20221441 and BK20241200. The authors would like to thank the support of Huawei Ascend Cloud Ecological Development Project.
\end{acks}
\clearpage

\begin{figure*}[p]
    \centering

    \includegraphics[
        page=2,
        width=\textwidth,
        height=0.925\textheight,
        keepaspectratio
    ]{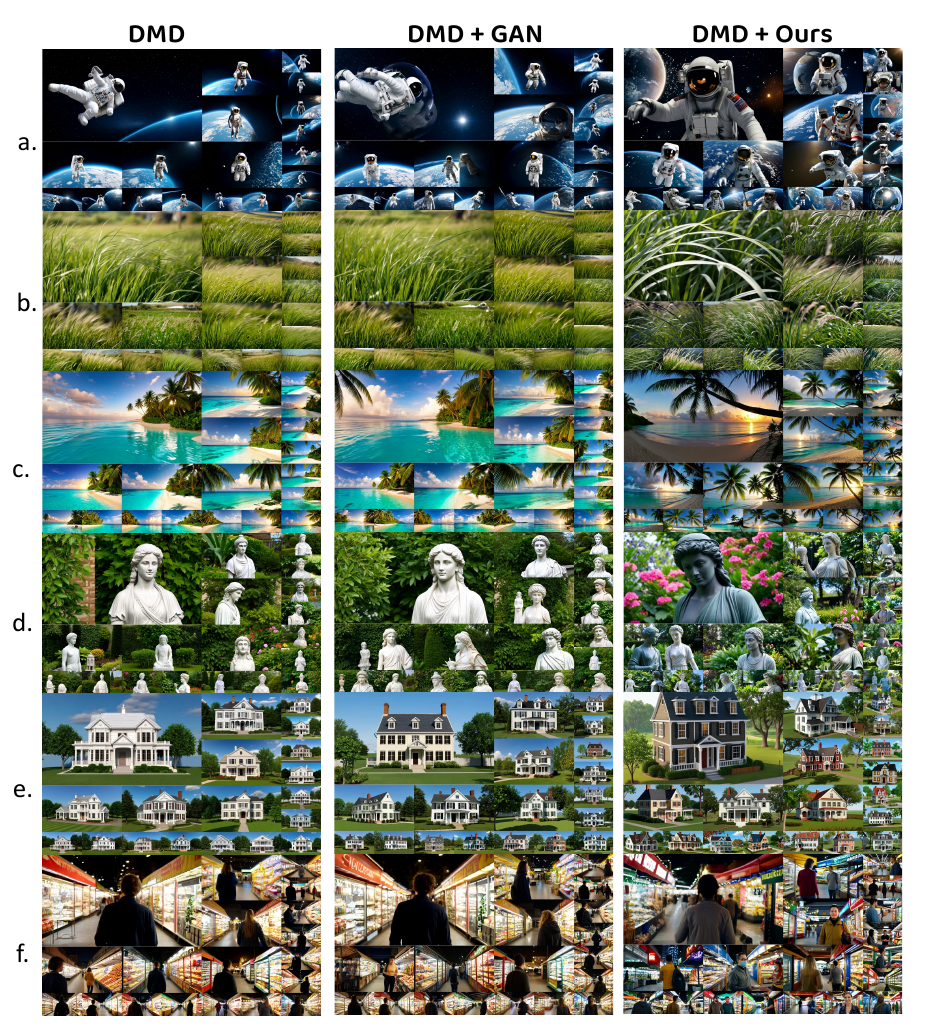}

    \caption{
    Diversity comparison results. 
    All methods are initialized from the same Self-Forcing ODE model. 
    For a fair comparison, we keep the random seeds identical across methods within each row, and generate 19 videos using seeds from 0 to 18 for each prompt. 
    The prompts are as follows:
    (a) \texttt{An astronaut flying in space, featuring a steady and smooth perspective};
    (b) \texttt{Grass swaying in the breeze};
    (c) \texttt{A tropical beach at sunrise, with palm trees and crystal-clear water in the foreground};
    (d) \texttt{A statue in a garden};
    (e) \texttt{A 3D model of a 1800s victorian house.};
    (f) \texttt{A person is walking through a night market}.
    }

    \label{fig:diversity_demo}
\end{figure*}
\clearpage

\clearpage

\begin{figure*}[p]
    \centering

    \includegraphics[
        page=1,
        width=\textwidth,
        height=\textheight,
        keepaspectratio
    ]{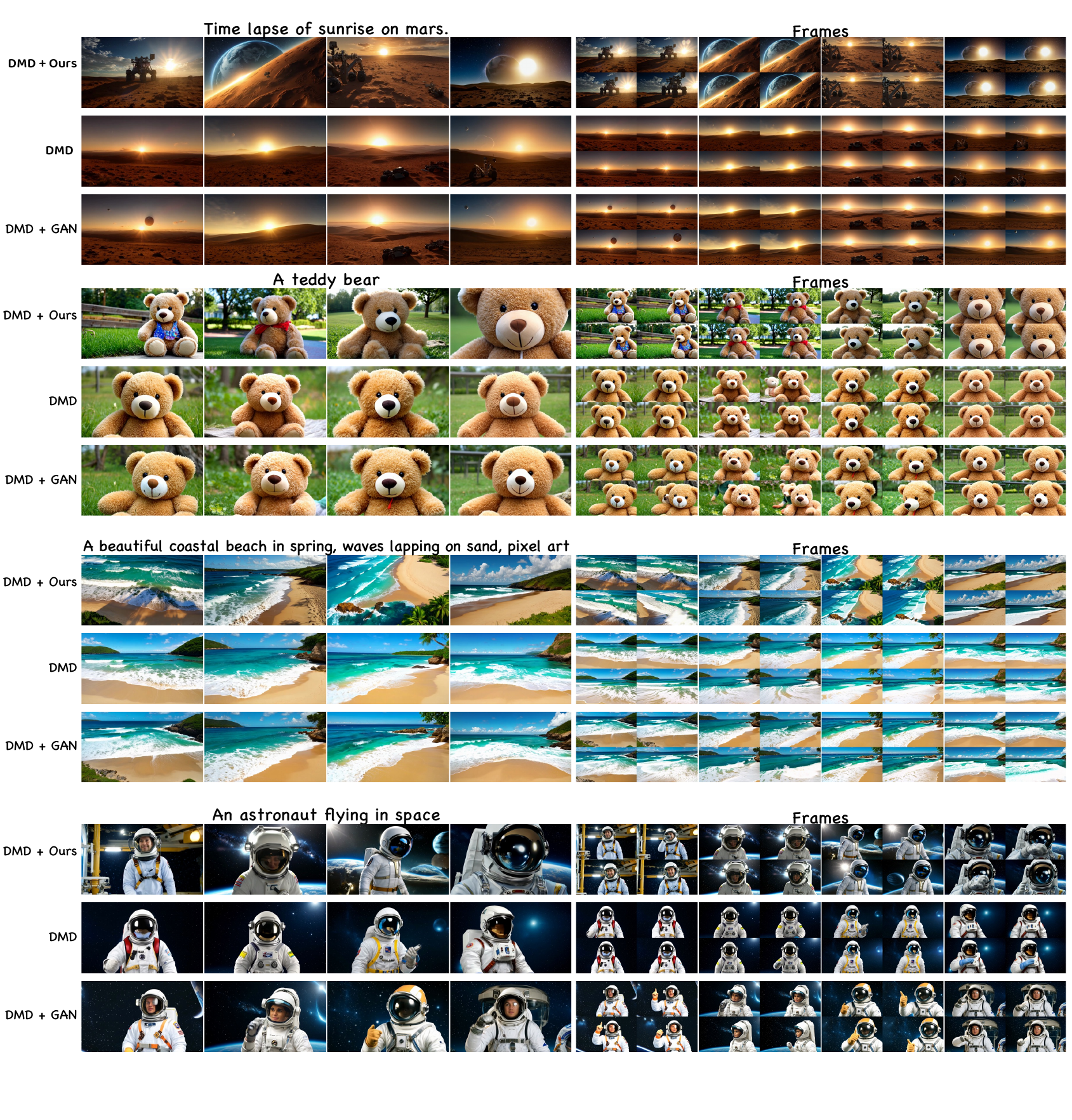}

    \caption{
    Diversity and temporal comparison results. 
    All methods are initialized from the same Self-Forcing ODE model. 
    For a fair comparison, we keep the random seeds identical across methods. 
    The prompt of each case is shown above the corresponding group. 
    For each prompt, the left side presents four videos generated with different seeds, and the right side shows the temporal evolution of these videos by uniformly sampling four frames from each corresponding 81-frame sequence. 
    }

    \label{fig:diversity_demo}
\end{figure*}
\clearpage

\bibliographystyle{ACM-Reference-Format}
\bibliography{sample-base}


\begin{thebibliography}{66}

%%% the Web address.

\ifx \showCODEN    \undefined \def \showCODEN     #1{\unskip}     \fi
\ifx \showISBNx    \undefined \def \showISBNx     #1{\unskip}     \fi
\ifx \showISBNxiii \undefined \def \showISBNxiii  #1{\unskip}     \fi
\ifx \showISSN     \undefined \def \showISSN      #1{\unskip}     \fi
\ifx \showLCCN     \undefined \def \showLCCN      #1{\unskip}     \fi
\ifx \shownote     \undefined \def \shownote      #1{#1}          \fi
\ifx \showarticletitle \undefined \def \showarticletitle #1{#1}   \fi
\ifx \showURL      \undefined \def \showURL       {\relax}        \fi
% invisible to TeX
\providecommand\bibfield[2]{#2}
\providecommand\bibinfo[2]{#2}
\providecommand\natexlab[1]{#1}
\providecommand\showeprint[2][]{arXiv:#2}

\bibitem[An et~al\mbox{.}(2026)]%
        {an2026ai}
\bibfield{author}{\bibinfo{person}{Hongjun An}, \bibinfo{person}{Wenhan Hu}, \bibinfo{person}{Sida Huang}, \bibinfo{person}{Siqi Huang}, \bibinfo{person}{Ruanjun Li}, \bibinfo{person}{Yuanzhi Liang}, \bibinfo{person}{Jiawei Shao}, \bibinfo{person}{Yiliang Song}, \bibinfo{person}{Zihan Wang}, \bibinfo{person}{Cheng Yuan}, {et~al\mbox{.}}} \bibinfo{year}{2026}\natexlab{}.
\newblock \showarticletitle{Ai flow: Perspectives, scenarios, and approaches}.
\newblock \bibinfo{journal}{\emph{Vicinagearth}} \bibinfo{volume}{3}, \bibinfo{number}{1} (\bibinfo{year}{2026}), \bibinfo{pages}{1}.
\newblock


\bibitem[Blattmann et~al\mbox{.}(2023)]%
        {blattmann2023align}
\bibfield{author}{\bibinfo{person}{Andreas Blattmann}, \bibinfo{person}{Robin Rombach}, \bibinfo{person}{Huan Ling}, \bibinfo{person}{Tim Dockhorn}, \bibinfo{person}{Seung~Wook Kim}, \bibinfo{person}{Sanja Fidler}, {and} \bibinfo{person}{Karsten Kreis}.} \bibinfo{year}{2023}\natexlab{}.
\newblock \showarticletitle{Align your latents: High-resolution video synthesis with latent diffusion models}. In \bibinfo{booktitle}{\emph{CVPR}}. \bibinfo{pages}{22563--22575}.
\newblock


\bibitem[Chen et~al\mbox{.}(2025b)]%
        {chen2025sana}
\bibfield{author}{\bibinfo{person}{Junsong Chen}, \bibinfo{person}{Shuchen Xue}, \bibinfo{person}{Yuyang Zhao}, \bibinfo{person}{Jincheng Yu}, \bibinfo{person}{Sayak Paul}, \bibinfo{person}{Junyu Chen}, \bibinfo{person}{Han Cai}, \bibinfo{person}{Song Han}, {and} \bibinfo{person}{Enze Xie}.} \bibinfo{year}{2025}\natexlab{b}.
\newblock \showarticletitle{Sana-sprint: One-step diffusion with continuous-time consistency distillation}. In \bibinfo{booktitle}{\emph{ICCV}}. \bibinfo{pages}{16185--16195}.
\newblock


\bibitem[Chen et~al\mbox{.}(2026a)]%
        {chen2026full4dgeneratingfullscope4d}
\bibfield{author}{\bibinfo{person}{Tingxi Chen}, \bibinfo{person}{Ke Hao}, \bibinfo{person}{Yabo Chen}, \bibinfo{person}{Zhengxue Cheng}, \bibinfo{person}{Rong Xie}, \bibinfo{person}{Li Song}, \bibinfo{person}{Haibin Huang}, \bibinfo{person}{Chi Zhang}, {and} \bibinfo{person}{Xuelong Li}.} \bibinfo{year}{2026}\natexlab{a}.
\newblock \bibinfo{title}{Full-4D: Generating Full-Scope 4D Scenes from a Single-View Video}.
\newblock
\showeprint[arxiv]{2605.25500}~[cs.CV]
\urldef\tempurl%
\url{https://arxiv.org/abs/2605.25500}
\showURL{%
\tempurl}


\bibitem[Chen et~al\mbox{.}(2026b)]%
        {chen2026generative}
\bibfield{author}{\bibinfo{person}{Xiangyu Chen}, \bibinfo{person}{Jixiang Luo}, \bibinfo{person}{Jingyu Xu}, \bibinfo{person}{Fangqiu Yi}, \bibinfo{person}{Chi Zhang}, {and} \bibinfo{person}{Xuelong Li}.} \bibinfo{year}{2026}\natexlab{b}.
\newblock \showarticletitle{Generative video compression: towards 0.01\% compression rate for video transmission}.
\newblock \bibinfo{journal}{\emph{Vicinagearth}} \bibinfo{volume}{3}, \bibinfo{number}{1} (\bibinfo{year}{2026}), \bibinfo{pages}{7}.
\newblock


\bibitem[Chen et~al\mbox{.}(2024a)]%
        {chen2024cascadezero123imagehighlyconsistent}
\bibfield{author}{\bibinfo{person}{Yabo Chen}, \bibinfo{person}{Jiemin Fang}, \bibinfo{person}{Yuyang Huang}, \bibinfo{person}{Taoran Yi}, \bibinfo{person}{Xiaopeng Zhang}, \bibinfo{person}{Lingxi Xie}, \bibinfo{person}{Xinggang Wang}, \bibinfo{person}{Wenrui Dai}, \bibinfo{person}{Hongkai Xiong}, {and} \bibinfo{person}{Qi Tian}.} \bibinfo{year}{2024}\natexlab{a}.
\newblock \bibinfo{title}{Cascade-Zero123: One Image to Highly Consistent 3D with Self-Prompted Nearby Views}.
\newblock
\showeprint[arxiv]{2312.04424}~[cs.CV]
\urldef\tempurl%
\url{https://arxiv.org/abs/2312.04424}
\showURL{%
\tempurl}


\bibitem[Chen et~al\mbox{.}(2025a)]%
        {chen2025teleworld}
\bibfield{author}{\bibinfo{person}{Yabo Chen}, \bibinfo{person}{Yuanzhi Liang}, \bibinfo{person}{Jiepeng Wang}, \bibinfo{person}{Tingxi Chen}, \bibinfo{person}{Junfei Cheng}, \bibinfo{person}{Zixiao Gu}, \bibinfo{person}{Yuyang Huang}, \bibinfo{person}{Zicheng Jiang}, \bibinfo{person}{Wei Li}, \bibinfo{person}{Tian Li}, {et~al\mbox{.}}} \bibinfo{year}{2025}\natexlab{a}.
\newblock \showarticletitle{TeleWorld: Towards Dynamic Multimodal Synthesis with a 4D World Model}.
\newblock \bibinfo{journal}{\emph{arXiv preprint arXiv:2601.00051}} (\bibinfo{year}{2025}).
\newblock


\bibitem[Chen et~al\mbox{.}(2024b)]%
        {chen2024liftimage3dliftingsingleimage}
\bibfield{author}{\bibinfo{person}{Yabo Chen}, \bibinfo{person}{Chen Yang}, \bibinfo{person}{Jiemin Fang}, \bibinfo{person}{Xiaopeng Zhang}, \bibinfo{person}{Lingxi Xie}, \bibinfo{person}{Wei Shen}, \bibinfo{person}{Wenrui Dai}, \bibinfo{person}{Hongkai Xiong}, {and} \bibinfo{person}{Qi Tian}.} \bibinfo{year}{2024}\natexlab{b}.
\newblock \bibinfo{title}{LiftImage3D: Lifting Any Single Image to 3D Gaussians with Video Generation Priors}.
\newblock
\showeprint[arxiv]{2412.09597}~[cs.CV]
\urldef\tempurl%
\url{https://arxiv.org/abs/2412.09597}
\showURL{%
\tempurl}


\bibitem[Cui et~al\mbox{.}(2026)]%
        {cui2026lol}
\bibfield{author}{\bibinfo{person}{Justin Cui}, \bibinfo{person}{Jie Wu}, \bibinfo{person}{Ming Li}, \bibinfo{person}{Tao Yang}, \bibinfo{person}{Xiaojie Li}, \bibinfo{person}{Rui Wang}, \bibinfo{person}{Andrew Bai}, \bibinfo{person}{Yuanhao Ban}, {and} \bibinfo{person}{Cho-Jui Hsieh}.} \bibinfo{year}{2026}\natexlab{}.
\newblock \showarticletitle{LoL: Longer than Longer, Scaling Video Generation to Hour}.
\newblock \bibinfo{journal}{\emph{arXiv preprint arXiv:2601.16914}} (\bibinfo{year}{2026}).
\newblock


\bibitem[Friedman and Dieng(2023)]%
        {friedman2023the}
\bibfield{author}{\bibinfo{person}{Dan Friedman} {and} \bibinfo{person}{Adji~Bousso Dieng}.} \bibinfo{year}{2023}\natexlab{}.
\newblock \showarticletitle{The Vendi Score: A Diversity Evaluation Metric for Machine Learning}.
\newblock \bibinfo{journal}{\emph{TMLR}} (\bibinfo{year}{2023}).
\newblock
\showISSN{2835-8856}


\bibitem[Fuest et~al\mbox{.}(2026)]%
        {fuest2026diffusion}
\bibfield{author}{\bibinfo{person}{Michael Fuest}, \bibinfo{person}{Pingchuan Ma}, \bibinfo{person}{Ming Gui}, \bibinfo{person}{Johannes Schusterbauer}, \bibinfo{person}{Vincent~Tao Hu}, {and} \bibinfo{person}{Bj{\"o}rn Ommer}.} \bibinfo{year}{2026}\natexlab{}.
\newblock \showarticletitle{Diffusion models and representation learning: A survey}.
\newblock \bibinfo{journal}{\emph{TPAMI}} (\bibinfo{year}{2026}).
\newblock


\bibitem[Gandikota and Bau(2026)]%
        {gandikota2026distilling}
\bibfield{author}{\bibinfo{person}{Rohit Gandikota} {and} \bibinfo{person}{David Bau}.} \bibinfo{year}{2026}\natexlab{}.
\newblock \showarticletitle{Distilling diversity and control in diffusion models}. In \bibinfo{booktitle}{\emph{WACV}}. \bibinfo{pages}{1304--1313}.
\newblock


\bibitem[Geng et~al\mbox{.}(2026)]%
        {geng2026mean}
\bibfield{author}{\bibinfo{person}{Zhengyang Geng}, \bibinfo{person}{Mingyang Deng}, \bibinfo{person}{Xingjian Bai}, \bibinfo{person}{Zico Kolter}, {and} \bibinfo{person}{Kaiming He}.} \bibinfo{year}{2026}\natexlab{}.
\newblock \showarticletitle{Mean flows for one-step generative modeling}.
\newblock \bibinfo{journal}{\emph{NeurIPS}}  \bibinfo{volume}{38} (\bibinfo{year}{2026}), \bibinfo{pages}{75460--75482}.
\newblock


\bibitem[Gu et~al\mbox{.}(2026)]%
        {gu2026searchtoworldevaluation3dworld}
\bibfield{author}{\bibinfo{person}{Zixiao Gu}, \bibinfo{person}{Yabo Chen}, \bibinfo{person}{Xunzhi Xiang}, \bibinfo{person}{Yu He}, \bibinfo{person}{Haibin Huang}, \bibinfo{person}{Chi Zhang}, \bibinfo{person}{Yunbo Wang}, {and} \bibinfo{person}{Xuelong Li}.} \bibinfo{year}{2026}\natexlab{}.
\newblock \bibinfo{title}{Search-to-World: Evaluation of 3D World Delivery from User Request through Web Search}.
\newblock
\showeprint[arxiv]{2609.07605}~[cs.CV]
\urldef\tempurl%
\url{https://arxiv.org/abs/2609.07605}
\showURL{%
\tempurl}


\bibitem[Hong et~al\mbox{.}(2023)]%
        {hong2023cogvideo}
\bibfield{author}{\bibinfo{person}{Wenyi Hong}, \bibinfo{person}{Ming Ding}, \bibinfo{person}{Wendi Zheng}, \bibinfo{person}{Xinghan Liu}, {and} \bibinfo{person}{Jie Tang}.} \bibinfo{year}{2023}\natexlab{}.
\newblock \showarticletitle{CogVideo: Large-scale Pretraining for Text-to-Video Generation via Transformers}. In \bibinfo{booktitle}{\emph{ICLR}}.
\newblock


\bibitem[Hu(2024)]%
        {hu2024animate}
\bibfield{author}{\bibinfo{person}{Li Hu}.} \bibinfo{year}{2024}\natexlab{}.
\newblock \showarticletitle{Animate anyone: Consistent and controllable image-to-video synthesis for character animation}. In \bibinfo{booktitle}{\emph{CVPR}}. \bibinfo{pages}{8153--8163}.
\newblock


\bibitem[Huang et~al\mbox{.}(2025b)]%
        {huang2025self}
\bibfield{author}{\bibinfo{person}{Xun Huang}, \bibinfo{person}{Zhengqi Li}, \bibinfo{person}{Guande He}, \bibinfo{person}{Mingyuan Zhou}, {and} \bibinfo{person}{Eli Shechtman}.} \bibinfo{year}{2025}\natexlab{b}.
\newblock \showarticletitle{Self forcing: Bridging the train-test gap in autoregressive video diffusion}.
\newblock \bibinfo{journal}{\emph{arXiv preprint arXiv:2506.08009}} (\bibinfo{year}{2025}).
\newblock


\bibitem[Huang et~al\mbox{.}(2026)]%
        {huang2026cineweavertrainingfreereferencecontrollablemultishot}
\bibfield{author}{\bibinfo{person}{Yuyang Huang}, \bibinfo{person}{Yabo Chen}, \bibinfo{person}{Wenrui Dai}, \bibinfo{person}{Ziyang Zheng}, \bibinfo{person}{Haibin Huang}, \bibinfo{person}{Chi Zhang}, \bibinfo{person}{Junni Zou}, \bibinfo{person}{Hongkai Xiong}, {and} \bibinfo{person}{Xuelong Li}.} \bibinfo{year}{2026}\natexlab{}.
\newblock \bibinfo{title}{CineWeaver: Training-Free Reference-Controllable Multi-Shot Long Video Generation for Cinematic Storytelling}.
\newblock
\showeprint[arxiv]{2607.26529}~[cs.CV]
\urldef\tempurl%
\url{https://arxiv.org/abs/2607.26529}
\showURL{%
\tempurl}


\bibitem[Huang et~al\mbox{.}(2025a)]%
        {huang2025zero}
\bibfield{author}{\bibinfo{person}{Yuyang Huang}, \bibinfo{person}{Yabo Chen}, \bibinfo{person}{Li Ding}, \bibinfo{person}{Xiaopeng Zhang}, \bibinfo{person}{Wenrui Dai}, \bibinfo{person}{Junni Zou}, \bibinfo{person}{Hongkai Xiong}, {and} \bibinfo{person}{Qi Tian}.} \bibinfo{year}{2025}\natexlab{a}.
\newblock \showarticletitle{Im-zero: Instance-level motion controllable video generation in a zero-shot manner}. In \bibinfo{booktitle}{\emph{2025 IEEE/CVF Conference on Computer Vision and Pattern Recognition (CVPR)}}. IEEE, \bibinfo{pages}{7265--7275}.
\newblock


\bibitem[Huang et~al\mbox{.}(2024a)]%
        {huang2024domainfusion}
\bibfield{author}{\bibinfo{person}{Yuyang Huang}, \bibinfo{person}{Yabo Chen}, \bibinfo{person}{Yuchen Liu}, \bibinfo{person}{Xiaopeng Zhang}, \bibinfo{person}{Wenrui Dai}, \bibinfo{person}{Hongkai Xiong}, {and} \bibinfo{person}{Qi Tian}.} \bibinfo{year}{2024}\natexlab{a}.
\newblock \showarticletitle{DomainFusion: Generalizing to unseen domains with latent diffusion models}. In \bibinfo{booktitle}{\emph{European conference on computer vision}}. Springer, \bibinfo{pages}{480--498}.
\newblock


\bibitem[Huang et~al\mbox{.}(2024b)]%
        {huang2024vbench}
\bibfield{author}{\bibinfo{person}{Ziqi Huang}, \bibinfo{person}{Yinan He}, \bibinfo{person}{Jiashuo Yu}, \bibinfo{person}{Fan Zhang}, \bibinfo{person}{Chenyang Si}, \bibinfo{person}{Yuming Jiang}, \bibinfo{person}{Yuanhan Zhang}, \bibinfo{person}{Tianxing Wu}, \bibinfo{person}{Qingyang Jin}, \bibinfo{person}{Nattapol Chanpaisit}, {et~al\mbox{.}}} \bibinfo{year}{2024}\natexlab{b}.
\newblock \showarticletitle{Vbench: Comprehensive benchmark suite for video generative models}. In \bibinfo{booktitle}{\emph{CVPR}}. \bibinfo{pages}{21807--21818}.
\newblock


\bibitem[Ibarrola and Grace(2024)]%
        {ibarrola2024measuring}
\bibfield{author}{\bibinfo{person}{Francisco Ibarrola} {and} \bibinfo{person}{Kazjon Grace}.} \bibinfo{year}{2024}\natexlab{}.
\newblock \showarticletitle{Measuring diversity in co-creative image generation}.
\newblock \bibinfo{journal}{\emph{arXiv preprint arXiv:2403.13826}} (\bibinfo{year}{2024}).
\newblock


\bibitem[Jia et~al\mbox{.}(2026)]%
        {jia2026moga}
\bibfield{author}{\bibinfo{person}{Weinan Jia}, \bibinfo{person}{Yuning Lu}, \bibinfo{person}{Mengqi Huang}, \bibinfo{person}{Hualiang Wang}, \bibinfo{person}{Binyuan Huang}, \bibinfo{person}{Nan Chen}, \bibinfo{person}{Mu Liu}, \bibinfo{person}{Jidong Jiang}, {and} \bibinfo{person}{Zhendong Mao}.} \bibinfo{year}{2026}\natexlab{}.
\newblock \showarticletitle{Mo{GA}: Mixture-of-Groups Attention for End-to-End Long Video Generation}. In \bibinfo{booktitle}{\emph{ICLR}}.
\newblock


\bibitem[Kazimi et~al\mbox{.}(2025)]%
        {kazimi2025diverse}
\bibfield{author}{\bibinfo{person}{Tahira Kazimi}, \bibinfo{person}{Connor Dunlop}, {and} \bibinfo{person}{Pinar Yanardag}.} \bibinfo{year}{2025}\natexlab{}.
\newblock \showarticletitle{Diverse Video Generation with Determinantal Point Process-Guided Policy Optimization}.
\newblock \bibinfo{journal}{\emph{arXiv preprint arXiv:2511.20647}} (\bibinfo{year}{2025}).
\newblock


\bibitem[Kong et~al\mbox{.}(2024)]%
        {kong2024hunyuanvideo}
\bibfield{author}{\bibinfo{person}{Weijie Kong}, \bibinfo{person}{Qi Tian}, \bibinfo{person}{Zijian Zhang}, \bibinfo{person}{Rox Min}, \bibinfo{person}{Zuozhuo Dai}, \bibinfo{person}{Jin Zhou}, \bibinfo{person}{Jiangfeng Xiong}, \bibinfo{person}{Xin Li}, \bibinfo{person}{Bo Wu}, \bibinfo{person}{Jianwei Zhang}, {et~al\mbox{.}}} \bibinfo{year}{2024}\natexlab{}.
\newblock \showarticletitle{Hunyuanvideo: A systematic framework for large video generative models}.
\newblock \bibinfo{journal}{\emph{arXiv preprint arXiv:2412.03603}} (\bibinfo{year}{2024}).
\newblock


\bibitem[Lin et~al\mbox{.}(2026)]%
        {lin2026autoregressive}
\bibfield{author}{\bibinfo{person}{Shanchuan Lin}, \bibinfo{person}{Ceyuan Yang}, \bibinfo{person}{Hao He}, \bibinfo{person}{Jianwen Jiang}, \bibinfo{person}{Yuxi Ren}, \bibinfo{person}{Xin Xia}, \bibinfo{person}{Yang Zhao}, \bibinfo{person}{Xuefeng Xiao}, {and} \bibinfo{person}{Lu Jiang}.} \bibinfo{year}{2026}\natexlab{}.
\newblock \showarticletitle{Autoregressive adversarial post-training for real-time interactive video generation}.
\newblock \bibinfo{journal}{\emph{NeurIPS}}  \bibinfo{volume}{38} (\bibinfo{year}{2026}), \bibinfo{pages}{41061--41086}.
\newblock


\bibitem[Liu et~al\mbox{.}(2026a)]%
        {liu2026decoupled}
\bibfield{author}{\bibinfo{person}{Dongyang Liu}, \bibinfo{person}{Peng Gao}, \bibinfo{person}{David Liu}, \bibinfo{person}{Ruoyi Du}, \bibinfo{person}{Zhen Li}, \bibinfo{person}{Qilong Wu}, \bibinfo{person}{Xin Jin}, \bibinfo{person}{Sihan Cao}, \bibinfo{person}{Shifeng Zhang}, \bibinfo{person}{Steven HOI}, {and} \bibinfo{person}{Hongsheng Li}.} \bibinfo{year}{2026}\natexlab{a}.
\newblock \showarticletitle{Decoupled {DMD}: {CFG} Augmentation as the Spear, Distribution Matching as the Shield}. In \bibinfo{booktitle}{\emph{ICLR}}.
\newblock


\bibitem[Liu et~al\mbox{.}(2026b)]%
        {liu2026rolling}
\bibfield{author}{\bibinfo{person}{Kunhao Liu}, \bibinfo{person}{Wenbo Hu}, \bibinfo{person}{Jiale Xu}, \bibinfo{person}{Ying Shan}, {and} \bibinfo{person}{Shijian Lu}.} \bibinfo{year}{2026}\natexlab{b}.
\newblock \showarticletitle{Rolling Forcing: Autoregressive Long Video Diffusion in Real Time}. In \bibinfo{booktitle}{\emph{ICLR}}.
\newblock


\bibitem[Lu et~al\mbox{.}(2025b)]%
        {lu2025law}
\bibfield{author}{\bibinfo{person}{Dakuan Lu}, \bibinfo{person}{Jiaqi Zhang}, \bibinfo{person}{Cheng Yuan}, \bibinfo{person}{Jiawei Shao}, {and} \bibinfo{person}{Xuelong Li}.} \bibinfo{year}{2025}\natexlab{b}.
\newblock \showarticletitle{The Law of Multi-Model Collaboration: Scaling Limits of Model Ensembling for Large Language Models}.
\newblock \bibinfo{journal}{\emph{arXiv preprint arXiv:2512.23340}} (\bibinfo{year}{2025}).
\newblock


\bibitem[Lu et~al\mbox{.}(2025a)]%
        {lu2025adversarial}
\bibfield{author}{\bibinfo{person}{Yanzuo Lu}, \bibinfo{person}{Yuxi Ren}, \bibinfo{person}{Xin Xia}, \bibinfo{person}{Shanchuan Lin}, \bibinfo{person}{Xing Wang}, \bibinfo{person}{Xuefeng Xiao}, \bibinfo{person}{Andy~J Ma}, \bibinfo{person}{Xiaohua Xie}, {and} \bibinfo{person}{Jian-Huang Lai}.} \bibinfo{year}{2025}\natexlab{a}.
\newblock \showarticletitle{Adversarial distribution matching for diffusion distillation towards efficient image and video synthesis}. In \bibinfo{booktitle}{\emph{2025 IEEE/CVF International Conference on Computer Vision (ICCV)}}. IEEE, \bibinfo{pages}{16818--16829}.
\newblock


\bibitem[Luo et~al\mbox{.}(2023)]%
        {luo2023diff}
\bibfield{author}{\bibinfo{person}{Weijian Luo}, \bibinfo{person}{Tianyang Hu}, \bibinfo{person}{Shifeng Zhang}, \bibinfo{person}{Jiacheng Sun}, \bibinfo{person}{Zhenguo Li}, {and} \bibinfo{person}{Zhihua Zhang}.} \bibinfo{year}{2023}\natexlab{}.
\newblock \showarticletitle{Diff-instruct: A universal approach for transferring knowledge from pre-trained diffusion models}.
\newblock \bibinfo{journal}{\emph{NeurIPS}}  \bibinfo{volume}{36} (\bibinfo{year}{2023}), \bibinfo{pages}{76525--76546}.
\newblock


\bibitem[Luo et~al\mbox{.}(2024)]%
        {luo2024one}
\bibfield{author}{\bibinfo{person}{Weijian Luo}, \bibinfo{person}{Zemin Huang}, \bibinfo{person}{Zhengyang Geng}, \bibinfo{person}{J~Zico Kolter}, {and} \bibinfo{person}{Guo-jun Qi}.} \bibinfo{year}{2024}\natexlab{}.
\newblock \showarticletitle{One-step diffusion distillation through score implicit matching}.
\newblock \bibinfo{journal}{\emph{NeurIPS}}  \bibinfo{volume}{37} (\bibinfo{year}{2024}), \bibinfo{pages}{115377--115408}.
\newblock


\bibitem[Nguyen and Tran(2024)]%
        {nguyen2024swiftbrush}
\bibfield{author}{\bibinfo{person}{Thuan~Hoang Nguyen} {and} \bibinfo{person}{Anh Tran}.} \bibinfo{year}{2024}\natexlab{}.
\newblock \showarticletitle{Swiftbrush: One-step text-to-image diffusion model with variational score distillation}. In \bibinfo{booktitle}{\emph{CVPR}}. \bibinfo{pages}{7807--7816}.
\newblock


\bibitem[Oquab et~al\mbox{.}(2024)]%
        {oquab2024dinov}
\bibfield{author}{\bibinfo{person}{Maxime Oquab}, \bibinfo{person}{Timoth{\'e}e Darcet}, \bibinfo{person}{Th{\'e}o Moutakanni}, \bibinfo{person}{Huy~V. Vo}, \bibinfo{person}{Marc Szafraniec}, \bibinfo{person}{Vasil Khalidov}, \bibinfo{person}{Pierre Fernandez}, \bibinfo{person}{Daniel HAZIZA}, \bibinfo{person}{Francisco Massa}, \bibinfo{person}{Alaaeldin El-Nouby}, \bibinfo{person}{Mido Assran}, \bibinfo{person}{Nicolas Ballas}, \bibinfo{person}{Wojciech Galuba}, \bibinfo{person}{Russell Howes}, \bibinfo{person}{Po-Yao Huang}, \bibinfo{person}{Shang-Wen Li}, \bibinfo{person}{Ishan Misra}, \bibinfo{person}{Michael Rabbat}, \bibinfo{person}{Vasu Sharma}, \bibinfo{person}{Gabriel Synnaeve}, \bibinfo{person}{Hu Xu}, \bibinfo{person}{Herve Jegou}, \bibinfo{person}{Julien Mairal}, \bibinfo{person}{Patrick Labatut}, \bibinfo{person}{Armand Joulin}, {and} \bibinfo{person}{Piotr Bojanowski}.} \bibinfo{year}{2024}\natexlab{}.
\newblock \showarticletitle{{DINO}v2: Learning Robust Visual Features without Supervision}.
\newblock \bibinfo{journal}{\emph{TMLR}} (\bibinfo{year}{2024}).
\newblock
\showISSN{2835-8856}
\newblock
\shownote{Featured Certification}.


\bibitem[Radford et~al\mbox{.}(2021)]%
        {radford2021learning}
\bibfield{author}{\bibinfo{person}{Alec Radford}, \bibinfo{person}{Jong~Wook Kim}, \bibinfo{person}{Chris Hallacy}, \bibinfo{person}{Aditya Ramesh}, \bibinfo{person}{Gabriel Goh}, \bibinfo{person}{Sandhini Agarwal}, \bibinfo{person}{Girish Sastry}, \bibinfo{person}{Amanda Askell}, \bibinfo{person}{Pamela Mishkin}, \bibinfo{person}{Jack Clark}, {et~al\mbox{.}}} \bibinfo{year}{2021}\natexlab{}.
\newblock \showarticletitle{Learning transferable visual models from natural language supervision}. In \bibinfo{booktitle}{\emph{ICML}}. PmLR, \bibinfo{pages}{8748--8763}.
\newblock


\bibitem[Sauer et~al\mbox{.}(2024)]%
        {sauer2024adversarial}
\bibfield{author}{\bibinfo{person}{Axel Sauer}, \bibinfo{person}{Dominik Lorenz}, \bibinfo{person}{Andreas Blattmann}, {and} \bibinfo{person}{Robin Rombach}.} \bibinfo{year}{2024}\natexlab{}.
\newblock \showarticletitle{Adversarial diffusion distillation}. In \bibinfo{booktitle}{\emph{ECCV}}. Springer, \bibinfo{pages}{87--103}.
\newblock


\bibitem[Shao and Li(2025)]%
        {shao2025ai}
\bibfield{author}{\bibinfo{person}{Jiawei Shao} {and} \bibinfo{person}{Xuelong Li}.} \bibinfo{year}{2025}\natexlab{}.
\newblock \showarticletitle{Ai flow at the network edge}.
\newblock \bibinfo{journal}{\emph{IEEE Network}} \bibinfo{volume}{40}, \bibinfo{number}{1} (\bibinfo{year}{2025}), \bibinfo{pages}{330--336}.
\newblock


\bibitem[Shen et~al\mbox{.}(2025)]%
        {shen2025efficient}
\bibfield{author}{\bibinfo{person}{Hui Shen}, \bibinfo{person}{Jingxuan Zhang}, \bibinfo{person}{Boning Xiong}, \bibinfo{person}{Rui Hu}, \bibinfo{person}{Shoufa Chen}, \bibinfo{person}{Zhongwei Wan}, \bibinfo{person}{Xin Wang}, \bibinfo{person}{Yu Zhang}, \bibinfo{person}{Zixuan Gong}, \bibinfo{person}{Guangyin Bao}, {et~al\mbox{.}}} \bibinfo{year}{2025}\natexlab{}.
\newblock \showarticletitle{Efficient diffusion models: A survey}.
\newblock \bibinfo{journal}{\emph{arXiv preprint arXiv:2502.06805}} (\bibinfo{year}{2025}).
\newblock


\bibitem[Song et~al\mbox{.}(2026)]%
        {song2026syncdit}
\bibfield{author}{\bibinfo{person}{Quanyue Song}, \bibinfo{person}{Zhizhi Guo}, \bibinfo{person}{Yishan He}, \bibinfo{person}{Zhihao Wang}, \bibinfo{person}{Zhixiang He}, \bibinfo{person}{Chi Zhang}, \bibinfo{person}{Caigui Jiang}, {and} \bibinfo{person}{Xuelong Li}.} \bibinfo{year}{2026}\natexlab{}.
\newblock \showarticletitle{SyncDIT: audio-visual aligned video generation with audio synchronization feature}.
\newblock \bibinfo{journal}{\emph{Vicinagearth}} \bibinfo{volume}{3}, \bibinfo{number}{1} (\bibinfo{year}{2026}), \bibinfo{pages}{5}.
\newblock


\bibitem[Song et~al\mbox{.}(2023)]%
        {song2023consistencymodels}
\bibfield{author}{\bibinfo{person}{Yang Song}, \bibinfo{person}{Prafulla Dhariwal}, \bibinfo{person}{Mark Chen}, {and} \bibinfo{person}{Ilya Sutskever}.} \bibinfo{year}{2023}\natexlab{}.
\newblock \bibinfo{title}{Consistency Models}.
\newblock
\showeprint[arxiv]{2303.01469}~[cs.LG]
\urldef\tempurl%
\url{https://arxiv.org/abs/2303.01469}
\showURL{%
\tempurl}


\bibitem[Sun et~al\mbox{.}(2025)]%
        {worldplay2025}
\bibfield{author}{\bibinfo{person}{Wenqiang Sun}, \bibinfo{person}{Haiyu Zhang}, \bibinfo{person}{Haoyuan Wang}, \bibinfo{person}{Junta Wu}, \bibinfo{person}{Zehan Wang}, \bibinfo{person}{Zhenwei Wang}, \bibinfo{person}{Yunhong Wang}, \bibinfo{person}{Jun Zhang}, \bibinfo{person}{Tengfei Wang}, {and} \bibinfo{person}{Chunchao Guo}.} \bibinfo{year}{2025}\natexlab{}.
\newblock \showarticletitle{WorldPlay: Towards Long-Term Geometric Consistency for Real-Time Interactive World Model}.
\newblock \bibinfo{journal}{\emph{arXiv preprint}} (\bibinfo{year}{2025}).
\newblock


\bibitem[Szegedy et~al\mbox{.}(2016)]%
        {szegedy2016rethinking}
\bibfield{author}{\bibinfo{person}{Christian Szegedy}, \bibinfo{person}{Vincent Vanhoucke}, \bibinfo{person}{Sergey Ioffe}, \bibinfo{person}{Jon Shlens}, {and} \bibinfo{person}{Zbigniew Wojna}.} \bibinfo{year}{2016}\natexlab{}.
\newblock \showarticletitle{Rethinking the inception architecture for computer vision}. In \bibinfo{booktitle}{\emph{Proceedings of the IEEE conference on computer vision and pattern recognition}}. \bibinfo{pages}{2818--2826}.
\newblock


\bibitem[Teng et~al\mbox{.}(2025)]%
        {teng2025magi}
\bibfield{author}{\bibinfo{person}{Hansi Teng}, \bibinfo{person}{Hongyu Jia}, \bibinfo{person}{Lei Sun}, \bibinfo{person}{Lingzhi Li}, \bibinfo{person}{Maolin Li}, \bibinfo{person}{Mingqiu Tang}, \bibinfo{person}{Shuai Han}, \bibinfo{person}{Tianning Zhang}, \bibinfo{person}{WQ Zhang}, \bibinfo{person}{Weifeng Luo}, {et~al\mbox{.}}} \bibinfo{year}{2025}\natexlab{}.
\newblock \showarticletitle{Magi-1: Autoregressive video generation at scale}.
\newblock \bibinfo{journal}{\emph{arXiv preprint arXiv:2505.13211}} (\bibinfo{year}{2025}).
\newblock


\bibitem[Wan et~al\mbox{.}(2025)]%
        {wan2025}
\bibfield{author}{\bibinfo{person}{Team Wan}, \bibinfo{person}{Ang Wang}, \bibinfo{person}{Baole Ai}, \bibinfo{person}{Bin Wen}, \bibinfo{person}{Chaojie Mao}, \bibinfo{person}{Chen-Wei Xie}, \bibinfo{person}{Di Chen}, \bibinfo{person}{Feiwu Yu}, \bibinfo{person}{Haiming Zhao}, \bibinfo{person}{Jianxiao Yang}, \bibinfo{person}{Jianyuan Zeng}, \bibinfo{person}{Jiayu Wang}, \bibinfo{person}{Jingfeng Zhang}, \bibinfo{person}{Jingren Zhou}, \bibinfo{person}{Jinkai Wang}, \bibinfo{person}{Jixuan Chen}, \bibinfo{person}{Kai Zhu}, \bibinfo{person}{Kang Zhao}, \bibinfo{person}{Keyu Yan}, \bibinfo{person}{Lianghua Huang}, \bibinfo{person}{Mengyang Feng}, \bibinfo{person}{Ningyi Zhang}, \bibinfo{person}{Pandeng Li}, \bibinfo{person}{Pingyu Wu}, \bibinfo{person}{Ruihang Chu}, \bibinfo{person}{Ruili Feng}, \bibinfo{person}{Shiwei Zhang}, \bibinfo{person}{Siyang Sun}, \bibinfo{person}{Tao Fang}, \bibinfo{person}{Tianxing Wang}, \bibinfo{person}{Tianyi Gui}, \bibinfo{person}{Tingyu Weng}, \bibinfo{person}{Tong Shen},
  \bibinfo{person}{Wei Lin}, \bibinfo{person}{Wei Wang}, \bibinfo{person}{Wei Wang}, \bibinfo{person}{Wenmeng Zhou}, \bibinfo{person}{Wente Wang}, \bibinfo{person}{Wenting Shen}, \bibinfo{person}{Wenyuan Yu}, \bibinfo{person}{Xianzhong Shi}, \bibinfo{person}{Xiaoming Huang}, \bibinfo{person}{Xin Xu}, \bibinfo{person}{Yan Kou}, \bibinfo{person}{Yangyu Lv}, \bibinfo{person}{Yifei Li}, \bibinfo{person}{Yijing Liu}, \bibinfo{person}{Yiming Wang}, \bibinfo{person}{Yingya Zhang}, \bibinfo{person}{Yitong Huang}, \bibinfo{person}{Yong Li}, \bibinfo{person}{You Wu}, \bibinfo{person}{Yu Liu}, \bibinfo{person}{Yulin Pan}, \bibinfo{person}{Yun Zheng}, \bibinfo{person}{Yuntao Hong}, \bibinfo{person}{Yupeng Shi}, \bibinfo{person}{Yutong Feng}, \bibinfo{person}{Zeyinzi Jiang}, \bibinfo{person}{Zhen Han}, \bibinfo{person}{Zhi-Fan Wu}, {and} \bibinfo{person}{Ziyu Liu}.} \bibinfo{year}{2025}\natexlab{}.
\newblock \showarticletitle{Wan: Open and Advanced Large-Scale Video Generative Models}.
\newblock \bibinfo{journal}{\emph{arXiv preprint arXiv:2503.20314}} (\bibinfo{year}{2025}).
\newblock


\bibitem[Wang et~al\mbox{.}(2026)]%
        {wang2026directingworldfastautoregressive}
\bibfield{author}{\bibinfo{person}{Haoyuan Wang}, \bibinfo{person}{Yabo Chen}, \bibinfo{person}{Haibin Huang}, \bibinfo{person}{Chi Zhang}, {and} \bibinfo{person}{Xuelong Li}.} \bibinfo{year}{2026}\natexlab{}.
\newblock \bibinfo{title}{Directing the World: Fast Autoregressive Video Generation with Compositional Human-Camera Control}.
\newblock
\showeprint[arxiv]{2606.27964}~[cs.CV]
\urldef\tempurl%
\url{https://arxiv.org/abs/2606.27964}
\showURL{%
\tempurl}


\bibitem[Wang and Yang(2024)]%
        {wang2024vidprom}
\bibfield{author}{\bibinfo{person}{Wenhao Wang} {and} \bibinfo{person}{Yi Yang}.} \bibinfo{year}{2024}\natexlab{}.
\newblock \showarticletitle{Vidprom: A million-scale real prompt-gallery dataset for text-to-video diffusion models}.
\newblock \bibinfo{journal}{\emph{NeurIPS}}  \bibinfo{volume}{37} (\bibinfo{year}{2024}), \bibinfo{pages}{65618--65642}.
\newblock


\bibitem[Wen et~al\mbox{.}(2025)]%
        {wen2025metricsolverslidinganchoredmetric}
\bibfield{author}{\bibinfo{person}{Tao Wen}, \bibinfo{person}{Jiepeng Wang}, \bibinfo{person}{Yabo Chen}, \bibinfo{person}{Shugong Xu}, \bibinfo{person}{Chi Zhang}, {and} \bibinfo{person}{Xuelong Li}.} \bibinfo{year}{2025}\natexlab{}.
\newblock \bibinfo{title}{Metric-Solver: Sliding Anchored Metric Depth Estimation from a Single Image}.
\newblock
\showeprint[arxiv]{2504.12103}~[cs.CV]
\urldef\tempurl%
\url{https://arxiv.org/abs/2504.12103}
\showURL{%
\tempurl}


\bibitem[Wu et~al\mbox{.}(2026)]%
        {wu2026diversity}
\bibfield{author}{\bibinfo{person}{Tianhe Wu}, \bibinfo{person}{Ruibin Li}, \bibinfo{person}{Lei Zhang}, {and} \bibinfo{person}{Kede Ma}.} \bibinfo{year}{2026}\natexlab{}.
\newblock \showarticletitle{Diversity-Preserved Distribution Matching Distillation for Fast Visual Synthesis}.
\newblock \bibinfo{journal}{\emph{arXiv preprint arXiv:2602.03139}} (\bibinfo{year}{2026}).
\newblock


\bibitem[Xiang et~al\mbox{.}(2025)]%
        {xiang2025macro}
\bibfield{author}{\bibinfo{person}{Xunzhi Xiang}, \bibinfo{person}{Yabo Chen}, \bibinfo{person}{Guiyu Zhang}, \bibinfo{person}{Zhongyu Wang}, \bibinfo{person}{Zhe Gao}, \bibinfo{person}{Quanming Xiang}, \bibinfo{person}{Gonghu Shang}, \bibinfo{person}{Junqi Liu}, \bibinfo{person}{Haibin Huang}, \bibinfo{person}{Yang Gao}, {et~al\mbox{.}}} \bibinfo{year}{2025}\natexlab{}.
\newblock \showarticletitle{Macro-from-micro planning for high-quality and parallelized autoregressive long video generation}.
\newblock \bibinfo{journal}{\emph{arXiv preprint arXiv:2508.03334}} (\bibinfo{year}{2025}).
\newblock


\bibitem[Xiang et~al\mbox{.}(2026a)]%
        {xiang2026videoweaveunlockinggeometricconsistency}
\bibfield{author}{\bibinfo{person}{Xunzhi Xiang}, \bibinfo{person}{Zixuan Duan}, \bibinfo{person}{Yabo Chen}, \bibinfo{person}{Zhengxuan Wei}, \bibinfo{person}{Guiyu Zhang}, \bibinfo{person}{Zixiao Gu}, \bibinfo{person}{Zhe Gao}, \bibinfo{person}{Haibin Huang}, \bibinfo{person}{Chi Zhang}, \bibinfo{person}{Qi Fan}, {and} \bibinfo{person}{Xuelong Li}.} \bibinfo{year}{2026}\natexlab{a}.
\newblock \bibinfo{title}{VideoWeave: Unlocking Geometric Consistency in Video Generation via Joint Geometry-Video Modeling}.
\newblock
\showeprint[arxiv]{2606.14162}~[cs.CV]
\urldef\tempurl%
\url{https://arxiv.org/abs/2606.14162}
\showURL{%
\tempurl}


\bibitem[Xiang et~al\mbox{.}(2026b)]%
        {xiang2026pathwise}
\bibfield{author}{\bibinfo{person}{Xunzhi Xiang}, \bibinfo{person}{Zixuan Duan}, \bibinfo{person}{Guiyu Zhang}, \bibinfo{person}{Haiyu Zhang}, \bibinfo{person}{Zhe Gao}, \bibinfo{person}{Junta Wu}, \bibinfo{person}{Shaofeng Zhang}, \bibinfo{person}{Tengfei Wang}, \bibinfo{person}{Qi Fan}, {and} \bibinfo{person}{Chunchao Guo}.} \bibinfo{year}{2026}\natexlab{b}.
\newblock \showarticletitle{Pathwise Test-Time Correction for Autoregressive Long Video Generation}.
\newblock \bibinfo{journal}{\emph{arXiv preprint arXiv:2602.05871}} (\bibinfo{year}{2026}).
\newblock


\bibitem[Xiang and Fan(2025)]%
        {xiang2025makeefficientdynamicsparse}
\bibfield{author}{\bibinfo{person}{Xunzhi Xiang} {and} \bibinfo{person}{Qi Fan}.} \bibinfo{year}{2025}\natexlab{}.
\newblock \bibinfo{title}{Make It Efficient: Dynamic Sparse Attention for Autoregressive Image Generation}.
\newblock
\showeprint[arxiv]{2506.18226}~[cs.CV]
\urldef\tempurl%
\url{https://arxiv.org/abs/2506.18226}
\showURL{%
\tempurl}


\bibitem[Xu et~al\mbox{.}(2025)]%
        {xu2025one}
\bibfield{author}{\bibinfo{person}{Yilun Xu}, \bibinfo{person}{Weili Nie}, {and} \bibinfo{person}{Arash Vahdat}.} \bibinfo{year}{2025}\natexlab{}.
\newblock \showarticletitle{One-step diffusion models with $ f $-divergence distribution matching}.
\newblock \bibinfo{journal}{\emph{arXiv preprint arXiv:2502.15681}} (\bibinfo{year}{2025}).
\newblock


\bibitem[Yang et~al\mbox{.}(2026)]%
        {yang2026longlive}
\bibfield{author}{\bibinfo{person}{Shuai Yang}, \bibinfo{person}{Wei Huang}, \bibinfo{person}{Ruihang Chu}, \bibinfo{person}{Yicheng Xiao}, \bibinfo{person}{Yuyang Zhao}, \bibinfo{person}{Xianbang Wang}, \bibinfo{person}{Muyang Li}, \bibinfo{person}{Enze Xie}, \bibinfo{person}{Ying-Cong Chen}, \bibinfo{person}{Yao Lu}, \bibinfo{person}{Song Han}, {and} \bibinfo{person}{Yukang Chen}.} \bibinfo{year}{2026}\natexlab{}.
\newblock \showarticletitle{LongLive: Real-time Interactive Long Video Generation}. In \bibinfo{booktitle}{\emph{ICLR}}.
\newblock


\bibitem[Yin et~al\mbox{.}(2024a)]%
        {yin2024improved}
\bibfield{author}{\bibinfo{person}{Tianwei Yin}, \bibinfo{person}{Micha{\"e}l Gharbi}, \bibinfo{person}{Taesung Park}, \bibinfo{person}{Richard Zhang}, \bibinfo{person}{Eli Shechtman}, \bibinfo{person}{Fredo Durand}, {and} \bibinfo{person}{William~T Freeman}.} \bibinfo{year}{2024}\natexlab{a}.
\newblock \showarticletitle{Improved distribution matching distillation for fast image synthesis}.
\newblock \bibinfo{journal}{\emph{NeurIPS}}  \bibinfo{volume}{37} (\bibinfo{year}{2024}), \bibinfo{pages}{47455--47487}.
\newblock


\bibitem[Yin et~al\mbox{.}(2024b)]%
        {yin2024one}
\bibfield{author}{\bibinfo{person}{Tianwei Yin}, \bibinfo{person}{Micha{\"e}l Gharbi}, \bibinfo{person}{Richard Zhang}, \bibinfo{person}{Eli Shechtman}, \bibinfo{person}{Fredo Durand}, \bibinfo{person}{William~T Freeman}, {and} \bibinfo{person}{Taesung Park}.} \bibinfo{year}{2024}\natexlab{b}.
\newblock \showarticletitle{One-step diffusion with distribution matching distillation}. In \bibinfo{booktitle}{\emph{CVPR}}. \bibinfo{pages}{6613--6623}.
\newblock


\bibitem[Yin et~al\mbox{.}(2025)]%
        {yin2025slow}
\bibfield{author}{\bibinfo{person}{Tianwei Yin}, \bibinfo{person}{Qiang Zhang}, \bibinfo{person}{Richard Zhang}, \bibinfo{person}{William~T Freeman}, \bibinfo{person}{Fredo Durand}, \bibinfo{person}{Eli Shechtman}, {and} \bibinfo{person}{Xun Huang}.} \bibinfo{year}{2025}\natexlab{}.
\newblock \showarticletitle{From slow bidirectional to fast autoregressive video diffusion models}. In \bibinfo{booktitle}{\emph{CVPR}}. \bibinfo{pages}{22963--22974}.
\newblock


\bibitem[Yu et~al\mbox{.}(2025)]%
        {yu2025context}
\bibfield{author}{\bibinfo{person}{Jiwen Yu}, \bibinfo{person}{Jianhong Bai}, \bibinfo{person}{Yiran Qin}, \bibinfo{person}{Quande Liu}, \bibinfo{person}{Xintao Wang}, \bibinfo{person}{Pengfei Wan}, \bibinfo{person}{Di Zhang}, {and} \bibinfo{person}{Xihui Liu}.} \bibinfo{year}{2025}\natexlab{}.
\newblock \showarticletitle{Context as memory: Scene-consistent interactive long video generation with memory retrieval}. In \bibinfo{booktitle}{\emph{Proceedings of the SIGGRAPH Asia 2025 Conference Papers}}. \bibinfo{pages}{1--11}.
\newblock


\bibitem[Yuan et~al\mbox{.}(2026a)]%
        {yuan2026enhancing}
\bibfield{author}{\bibinfo{person}{Cheng Yuan}, \bibinfo{person}{Zhenyu Jia}, \bibinfo{person}{Jiawei Shao}, {and} \bibinfo{person}{Xuelong Li}.} \bibinfo{year}{2026}\natexlab{a}.
\newblock \showarticletitle{Enhancing Neural Video Compression of Static Scenes with Positive-Incentive Noise}.
\newblock \bibinfo{journal}{\emph{arXiv preprint arXiv:2603.06095}} (\bibinfo{year}{2026}).
\newblock


\bibitem[Yuan et~al\mbox{.}(2026b)]%
        {helios}
\bibfield{author}{\bibinfo{person}{Shenghai Yuan}, \bibinfo{person}{Yuanyang Yin}, \bibinfo{person}{Zongjian Li}, \bibinfo{person}{Xinwei Huang}, \bibinfo{person}{Xiao Yang}, {and} \bibinfo{person}{Li Yuan}.} \bibinfo{year}{2026}\natexlab{b}.
\newblock \showarticletitle{Helios: Real Real-Time Long Video Generation Model}.
\newblock \bibinfo{journal}{\emph{arXiv preprint arXiv:2603.04379}} (\bibinfo{year}{2026}).
\newblock


\bibitem[Zang et~al\mbox{.}(2026)]%
        {zang2026instruction}
\bibfield{author}{\bibinfo{person}{Xianghao Zang}, \bibinfo{person}{Zijian Jiang}, \bibinfo{person}{Jiarong Cheng}, \bibinfo{person}{Qianrui Teng}, \bibinfo{person}{Ying He}, \bibinfo{person}{Yuxuan Mu}, \bibinfo{person}{Chao Ban}, \bibinfo{person}{Huayu Zhang}, \bibinfo{person}{Lanxiang Zhou}, \bibinfo{person}{Zerun Feng}, {et~al\mbox{.}}} \bibinfo{year}{2026}\natexlab{}.
\newblock \showarticletitle{Instruction-based image editing: a survey on data, models, evaluation, and applications}.
\newblock \bibinfo{journal}{\emph{Vicinagearth}} \bibinfo{volume}{3}, \bibinfo{number}{1} (\bibinfo{year}{2026}), \bibinfo{pages}{3}.
\newblock


\bibitem[Zhang et~al\mbox{.}(2026c)]%
        {Zhang_2026_CVPR}
\bibfield{author}{\bibinfo{person}{Guiyu Zhang}, \bibinfo{person}{Yabo Chen}, \bibinfo{person}{Xunzhi Xiang}, \bibinfo{person}{Junchao Huang}, \bibinfo{person}{Zhongyu Wang}, {and} \bibinfo{person}{Li Jiang}.} \bibinfo{year}{2026}\natexlab{c}.
\newblock \showarticletitle{SymphoMotion: Joint Control of Camera Motion and Object Dynamics for Coherent Video Generation}. In \bibinfo{booktitle}{\emph{Proceedings of the IEEE/CVF Conference on Computer Vision and Pattern Recognition (CVPR)}}. \bibinfo{pages}{11127--11137}.
\newblock


\bibitem[Zhang et~al\mbox{.}(2026a)]%
        {zhang2026tourphysicsbringingphysicsworld}
\bibfield{author}{\bibinfo{person}{Xin Zhang}, \bibinfo{person}{Yabo Chen}, \bibinfo{person}{Zixuan Duan}, \bibinfo{person}{Haibin Huang}, \bibinfo{person}{Chi Zhang}, \bibinfo{person}{Feng Xu}, {and} \bibinfo{person}{Xuelong Li}.} \bibinfo{year}{2026}\natexlab{a}.
\newblock \bibinfo{title}{TourPhysics: Bringing Physics to World Models for Exploration and Manipulation from a Single Image}.
\newblock
\showeprint[arxiv]{2609.04911}~[cs.CV]
\urldef\tempurl%
\url{https://arxiv.org/abs/2609.04911}
\showURL{%
\tempurl}


\bibitem[Zhang et~al\mbox{.}(2026b)]%
        {zhang2026physomniphysicsgroundedmultiobjectscene}
\bibfield{author}{\bibinfo{person}{Xin Zhang}, \bibinfo{person}{Yabo Chen}, \bibinfo{person}{Yijie Fang}, \bibinfo{person}{Wanying Qu}, \bibinfo{person}{Haibin Huang}, \bibinfo{person}{Chi Zhang}, \bibinfo{person}{Feng Xu}, {and} \bibinfo{person}{Xuelong Li}.} \bibinfo{year}{2026}\natexlab{b}.
\newblock \bibinfo{title}{PhysOmni: Physics-Grounded Multi-Object Scene Generation from a Single Image with Real-Time Interaction}.
\newblock
\showeprint[arxiv]{2605.20290}~[cs.GR]
\urldef\tempurl%
\url{https://arxiv.org/abs/2605.20290}
\showURL{%
\tempurl}


\bibitem[Zhou et~al\mbox{.}(2025)]%
        {zhou2025adversarial}
\bibfield{author}{\bibinfo{person}{Mingyuan Zhou}, \bibinfo{person}{Huangjie Zheng}, \bibinfo{person}{Yi Gu}, \bibinfo{person}{Zhendong Wang}, {and} \bibinfo{person}{Hai Huang}.} \bibinfo{year}{2025}\natexlab{}.
\newblock \showarticletitle{Adversarial Score identity Distillation: Rapidly Surpassing the Teacher in One Step}. In \bibinfo{booktitle}{\emph{ICLR}}.
\newblock


\bibitem[Zhu et~al\mbox{.}(2026)]%
        {zhu2026causal}
\bibfield{author}{\bibinfo{person}{Hongzhou Zhu}, \bibinfo{person}{Min Zhao}, \bibinfo{person}{Guande He}, \bibinfo{person}{Hang Su}, \bibinfo{person}{Chongxuan Li}, {and} \bibinfo{person}{Jun Zhu}.} \bibinfo{year}{2026}\natexlab{}.
\newblock \showarticletitle{Causal Forcing: Autoregressive Diffusion Distillation Done Right for High-Quality Real-Time Interactive Video Generation}.
\newblock \bibinfo{journal}{\emph{arXiv preprint arXiv:2602.02214}} (\bibinfo{year}{2026}).
\newblock


\end{thebibliography}

\end{document}